\documentclass[letterpaper]{article}
\usepackage{aaai19}

\usepackage{times}
\usepackage{helvet}
\usepackage{courier}
\usepackage{url}
\usepackage{enumitem}
\usepackage{siunitx}
\usepackage{physics}
\usepackage{amssymb}
\usepackage{amsmath}
\usepackage[makeroom]{cancel}
\usepackage{mathrsfs}
\usepackage{amsthm}

\usepackage{bm}

\usepackage{graphicx}
\usepackage[skip=0pt,labelformat=simple]{subcaption}
\DeclareCaptionFont{tiny}{\tiny}
\usepackage{multirow}
\usepackage[table,xcdraw]{xcolor}

\usepackage[linesnumbered,vlined,ruled]{algorithm2e}
\makeatletter
\newcommand\notsotiny{\@setfontsize\notsotiny{7}{8}}
\makeatother
\SetKwProg{Fn}{Function}{}{}
\SetAlFnt{\scriptsize}
\SetAlCapFnt{\normalfont\scriptsize}
\SetAlCapNameFnt{\scriptsize}
\SetArgSty{textrm}
\SetInd{0.1em}{0.5em}
\SetCommentSty{emph}
\usepackage{setspace}

\usepackage{tabularx, booktabs}
\usepackage{varioref}
\usepackage[noabbrev,capitalise]{cleveref}

\theoremstyle{definition}

\newtheorem{thm}{Theorem}

\newtheorem{cor}[thm]{Corollary}

\theoremstyle{definition}

\newtheorem{defn}{Definition}%[section]
\usepackage{thmtools}

\declaretheoremstyle[%
  spaceabove=0pt,%reduce or increase between theorem and proof
  spacebelow=0pt,%reduce or increase
  headfont=\normalfont\itshape,%
  postheadspace=1em,%
  qed=\qedsymbol%
]{mystyle}
\declaretheorem[name={Proof},style=mystyle,unnumbered,
]{pf}

\renewcommand\arraystretch{0.5}

\graphicspath {{figures/}}
\allowdisplaybreaks

\newcommand{\OPEN} {{\textsc{OPEN}}}
\newcommand{\STACK} {{\textsc{STACK}}}

\newcommand{\LIST} {{\textsc{LIST}}}
\newcommand{\PREC} {\boldsymbol{\pmb\prec}}
\providecommand{\agent}[1]{a_{#1}} %Represents a specified robot
\newcommand{\ignore}[1]{}

\begin{document}

\title{Searching with Consistent Prioritization for Multi-Agent Path Finding\thanks{The research at the University of Southern California was supported by the National Science Foundation (NSF) under grant numbers 1409987, 1724392, 1817189 and 1837779 as well as a gift from Amazon. The views and conclusions contained in this document are those of the authors and should not be interpreted as representing the official policies, either expressed or implied, of the sponsoring organizations, agencies or the U.S. government.}
}
\author{Hang Ma\textsuperscript{1},
Daniel Harabor\textsuperscript{2},
Peter J. Stuckey\textsuperscript{2},
Jiaoyang Li\textsuperscript{1},
Sven Koenig\textsuperscript{1}\\
\textsuperscript{1}University of Southern California\\
\textsuperscript{2}Monash University\\
hangma@usc.edu,
\{daniel.harabor,peter.stuckey\}@monash.edu,
\{jiaoyanl,skoenig\}@usc.edu
}

\maketitle

\begin{abstract}
We study prioritized planning for Multi-Agent Path Finding (MAPF). Existing prioritized MAPF algorithms depend on rule-of-thumb heuristics and random assignment to determine a fixed total priority ordering of all agents a priori. We instead explore the space of all possible partial priority orderings as part of a novel systematic and conflict-driven combinatorial search framework. In a variety of empirical comparisons, we demonstrate state-of-the-art solution qualities and success rates, often with similar runtimes to existing algorithms. We also develop new theoretical results that explore the limitations of prioritized planning, in terms of completeness and optimality, for the first time.
\end{abstract}

\section{Introduction}
Multi-Agent Path Finding (MAPF) is a coordination problem that arises in many applications, such as for aircraft-towing vehicles \cite{airporttug16}, warehouse and office robots \cite{kiva,DBLP:conf/ijcai/VelosoBCR15}, game characters \cite{MaAIIDE17}, and other multi-agent systems \cite{MaIEEE17}. The problem is to plan collision-free paths for multiple agents on a given graph from their given start vertices to their given target vertices \cite{MaAIMATTERS17}. The quality of a solution is measured by the flowtime (the sum of the arrival times of all agents at their target vertices) or the makespan (the maximum of the arrival times of all agents at their target vertices). MAPF is NP-hard to solve optimally \cite{YuLav13AAAI,MaAAAI16}. It can be solved with reductions to other well-studied combinatorial problems
\cite{Surynek15,YuLav13ICRA,erdem2013general} and dedicated MAPF algorithms
\cite{ODA11,PushAndSwap,EPEJAIR,DBLP:journals/ai/SharonSGF13,MStar,DBLP:journals/ai/SharonSFS15}, as described in several surveys \cite{MaWOMPF16,SoCS2017Surv}.

Prioritized MAPF algorithms~\cite{WHCA,WHCA06} are among the most efficient ones for solving MAPF.
They are based on the following simple prioritized-planning scheme \cite{ErdmannL87}: Each agent is given a unique priority and computes, in priority order, a minimum-cost path from its start vertex to its target vertex that does not collide with the (already planned) paths of all agents with higher priorities. Existing (standard) prioritized MAPF algorithms are well known for their small runtimes and are often used as parts of MAPF solvers~\cite{velagapudi2010decentralized,WangB11,CapVK15}. However, they determine a predefined total priority ordering of the agents a priori and can thus result in solutions of bad quality or even fail to find any solutions for solvable MAPF instances, where a different total priority ordering could have resulted in solutions of higher quality.

In this paper, we thus generalize the notion of prioritized planning from planning with a fixed total priority ordering on the agents to planning with all possible total priority orderings. Theoretically, we establish, for the first time, a conceptual framework for discussing the limitations of prioritized planning. For example, we identify the set of solutions resulting from prioritized MAPF algorithms and characterize classes of MAPF instances with respect to the completeness and optimality guarantees for different priority orderings.
We also develop two prioritized MAPF algorithms that improve the practicality of prioritized planning by systematically exploring ``good'' priority orderings. Our first algorithm, Conflict-Based Search with Priorities (CBSw/P), is an adaptation of Conflict-Based Search (CBS) \cite{DBLP:journals/ai/SharonSFS15}. It explores the space of all total priority orderings lazily using a systematic best-first search and introduces an ordered pair of agents only when their paths collide. Our second algorithm, Priority-Based Search (PBS), explores the space of all total priority orderings lazily using a systematic depth-first search. It can take a user-specified partial priority ordering as input, dynamically adds new ordered pairs of agents to it, and plans paths that are consistent with the resulting partial priority ordering. We show that standard prioritized MAPF algorithms are a special case of PBS.
Empirically, we evaluate our prioritized MAPF algorithms on a large number of MAPF instances. We find that CBSw/P often computes optimal or near-optimal solutions and is more efficient than a state-of-the-art version of CBS~\cite{FelnerICAPS18}. PBS also computes near-optimal solutions and is much more efficient than CBSw/P.
Moreover, PBS remains near-optimal and efficient for MAPF instances with more
than one hundred agents, finds solutions for many MAPF instances where
standard prioritized MAPF algorithms cannot, and solves well-formed MAPF instances with six hundred agents in less than a minute.

\section{Problem Definition}

We formalize MAPF as follows: We are given a connected undirected graph $G = (V,E)$ and $M$ agents $\{a_i~|~i\in[M]\}$ ($[M] = \{1,\ldots, M\}$). Each agent $a_i$ has a unique start vertex $s_i\in V$ and a unique target vertex $t_i\in V$. At each discrete time $t= 0, \ldots, \infty$, each agent either moves to an adjacent vertex or waits at the same vertex. Let $\pi_i(t)$ be the vertex occupied by agent $a_i$ at time $t$. A \emph{plan} consists of a set of paths, one path $\pi_i = \langle \pi_i(0), \ldots, \pi_i(T_i), \pi_i(T_i + 1), \ldots\rangle$ for each agent $a_i$, where $\pi_i(0) = s_i$ and $\pi_i(t) = t_i$ for all times $t = T_i, \ldots, \infty$. Specifically, the \emph{arrival time} $T_i$ of agent $a_i$ at its target vertex is defined to be the earliest time when it has reached its target vertex and stops moving. A \emph{vertex collision} is a tuple $\langle a_i,a_j,v,t \rangle$ where agents $a_i$ and $a_j$ occupy the same vertex $v$ at the same time $t$. An \emph{edge collision} is a tuple $\langle a_i,a_j,u,v,t \rangle$ where agents $a_i$ and $a_j$ traverse the same edge $(u,v)$ in opposite directions at the same time $t$. A solution is a plan that consists of collision-free paths for all agents. Its quality is measured by the \emph{flowtime} $\sum_{i\in[M]}T_i$, defined to be the sum of the arrival times of all agents.

\section{Prioritized Planning}

Prioritized planning~\cite{ErdmannL87} is a decoupled approach for
MAPF where agents are ordered by importance according to a predefined total priority ordering.
%The idea is simple: compute the set of paths one at a time in decreasing order
%of priority.
The idea is simple: One can plan for each agent individually rather than having to compute a plan for all agents simultaneously (as is the case for coupled MAPF algorithms).
This means that one plans for the highest priority agent first and computes its individually optimal path; i.e. it avoids only the fixed obstacles. One then plans for lower and lower priority agents and computes for each agent its individually optimal paths that avoids not only the fixed obstacles but also collisions with the (already planned) paths of all higher priority agents (treated as dynamic obstacles).
We generalize prioritized planning for MAPF by using partial priority ordering.
\begin{defn}
  A \emph{priority ordering} $\PREC$ is a strict partial order on $[M]$. Agent $a_i$ is of higher priority than agent $a_j$ iff $i\prec j$.
 \label{defn:priority}
\end{defn}
It does not offer completeness or optimality guarantees.
It is nevertheless popular because of its efficiency. Its main challenge is to determine a good priority ordering $\PREC$ since a bad one can result in solutions of low quality or even failures in solving the problem.
%Research on this topic can be broadly categorized as follows:
% \begin{itemize}[nosep]
%     \item
Global orderings assign fixed priorities to all agents a priori and resolve all collisions before movement begins. Priorities can be assigned arbitrarily~\cite{warren1990multiple,bennewitz2002finding,WHCA} or derived from the problem at hand~\cite{ErdmannL87}. For example, priorities can also be computed using heuristics such as the distances to their target vertices~\cite{van2005prioritized} or by preferring certain types of paths over others~\cite{buckley1989fast,ferrari1998multirobot}.
	%and by preferring agents with certain physical
	%characteristics~\cite{citation-goes-here}.
    %\item
Local orderings assign temporary priorities to some agents in order to resolve collisions on-the-fly. Such algorithms require agents to follow their assigned paths and, when an impasse is reached, priorities are assigned dynamically to determine who waits~\cite{ODonnellL89,azarm1997conflict}.
% \end{itemize}
Some existing works attempt to reason over the space of all total priority orderings, which is intractable in general as there are $M!$ permutations. \citeauthor{bennewitz2002finding} (\citeyear{bennewitz2002finding}) explore some of this space by generating several total priority orderings randomly as part of a hill-climbing scheme. \citeauthor{azarm1997conflict} (\citeyear{azarm1997conflict}) enumerate all total priority orderings for up to three agents.

%In (standard) prioritized planning is the name given to , we are given a pre-defined total priority ordering $\PREC$ on $[M]$. One of the standard prioritized MAPF algorithms is Cooperative A*(CA*) \cite{WHCA}. CA* plans paths for all single agents independently from the agent with minimal (highest) priority to the agent with maximal (lowest) priority. For an agent $a_j$, it uses a space-time A* to plan a time-optimal path from its start vertex $s_j$ to its goal vertex $g_j$ that does not collide with any agent $a_i$ with higher priority, namely $i \prec j$. In this paper, we generalize the pre-defined total priority ordering $\PREC$ to a partial priority ordering.

%In this section, we present formal definitions of Multi-Agent Path Finding
%In this section we give an overview of prioritized planning for multi-agent
%pathfinding and a generalised formulation
%and we generalise this idea from
%over the set of agents to
%concept of prioritized planning. We then show that
%prioritized planning with all possible priority orderings improves the
%solvability of the standard prioritized MAPF algorithm but is still incomplete
%and non-optimal.

\section{Theoretical Results}

\begin{figure}
  \centering
  \begin{minipage}[t]{0.48\columnwidth}
  \centering
  \hspace{-1em}
  \includegraphics[width=.8\textwidth]{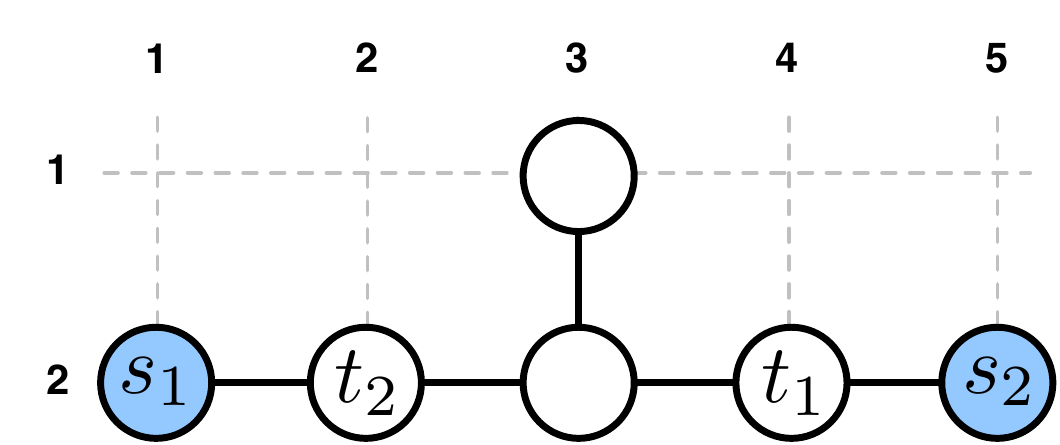}
  \caption{This MAPF instance is not solvable with any fixed priority ordering.}
 \label{fig:incomplete}
  \end{minipage}
  \hfill
  \begin{minipage}[t!]{0.48\columnwidth}
  \centering
  \includegraphics[width=.8\columnwidth]{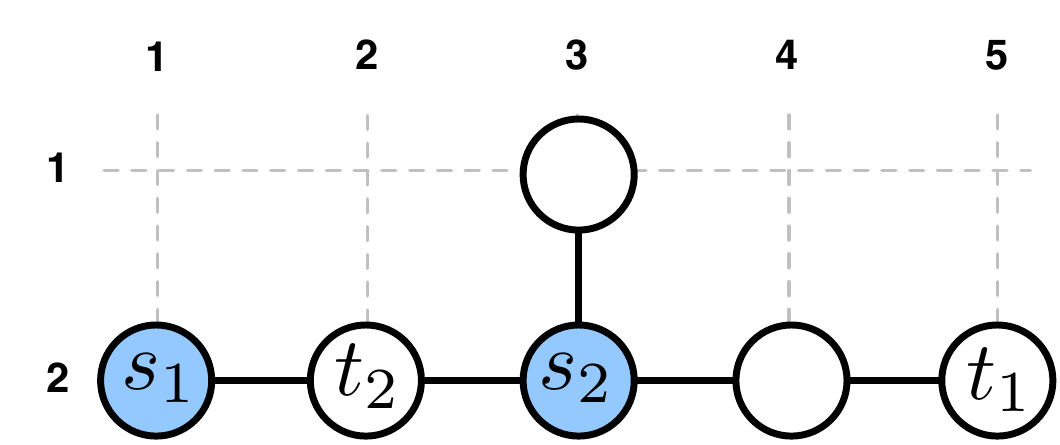}
  \caption{This MAPF instance is P-solvable and can be solved only with priority ordering $\lbrace 1 \prec 2 \rbrace$.}
 \label{fig:p_solvable}
  \end{minipage}
  \hfill
  \begin{minipage}[t]{0.48\columnwidth}
  \centering
  \hspace{-1em}
  \includegraphics[width=.8\columnwidth]{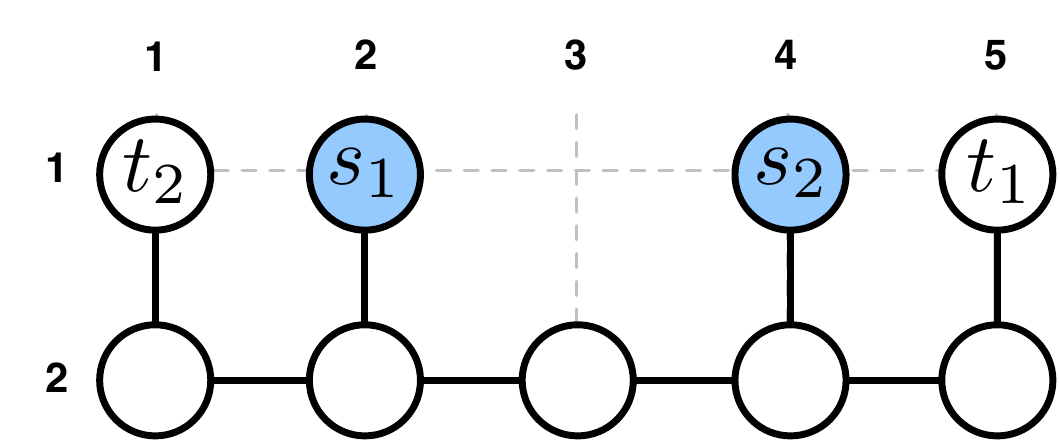}
  \caption{This MAPF instance is \emph{well-formed} and P-solvable for any total priority ordering.}
 \label{fig:wellformed}
  \end{minipage}
  \hfill
  \begin{minipage}[t]{0.48\columnwidth}
  \centering
  \hspace{-1em}
  \includegraphics[width=.8\columnwidth]{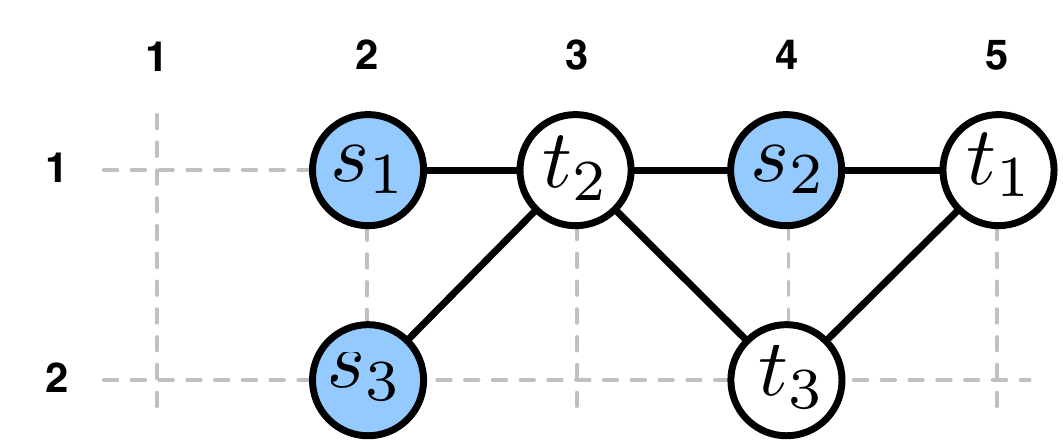}
  \caption{
      This MAPF instance is OP-solvable but prioritized planning is not guaranteed to find a solution.
  }
 \label{fig:op_solvable}
  \end{minipage}
  %\caption{Left: A MAPF instance with two agents $a_1$ and $a_2$. Right: A MAPF instance on a different graph.}\label{fig:instance}
\end{figure}

%\ddh{This section is under construction. Might not make sense yet}
We now analyze the effectiveness of prioritized planning,
in terms of completeness and optimality, on different classes of MAPF
instances. We first generalize a well known result \cite{ErdmannL87} to priority planning with a partial priority ordering.

%:We recap a well known incompleteness result and then we identify several
%:classes of MAPF problems for which this approach is provably complete.
%:We also give two new negative results showing that, even when a solution
%:can be found, there exists no fixed priority ordering which guarantees
%:optimality under two popular objective functions: sum-of-costs and makespan.
%:Finally, we characterise a new class of MAPF problems for which prioritized
%:planning is both complete and optimal.

%In particular, the ordered pair $i \prec j$ specifies that a time-minimal path for agent $a_i$ (with higher priority) is planned before a time-minimal path is planned for agent $a_j$ (with lower priority) that does not collide with the path for agent $a_i$.

%\begin{defn}
%Two priority ordering sets $(\emph{P}, \prec)$ and $(\emph{P}, \prec')$ (on the same priority set $[M]$) are \emph{consistent}, if and only if, for all  $i, j \in [M]$, $i \prec j$ implies $j \nprec' i$ and $i \prec' j$ implies $i \nprec j$.
%\end{defn}

\begin{thm}
%\cite{ErdmannL87}
Prioritized planning with an arbitrary priority ordering $\PREC$ is
incomplete for MAPF in general.
\label{thm:incomplete}
\end{thm}
\begin{pf}
The only three possible priority orderings for the counter-example shown in Figure~\ref{fig:incomplete} are $\lbrace 1 \prec 2 \rbrace$, $\lbrace 2 \prec 1 \rbrace$
or $\emptyset$. Prioritized planning for none of them results in a solution.
\end{pf}
%Next, we show that there exists a large class of practical MAPF instances,
%which we call \emph{P-solvable}, where prioritized planning can be used to
%find a solution.

Next, we define a class of MAPF instances which we call \emph{P-solvable}.
A MAPF instance is in this class iff it has a solution that can be computed with prioritized planning, that is, a fixed priority ordering exists where higher priority agents never wait for lower priority agents.

\begin{defn}
A solution $L = \lbrace \pi_i~|~i \in [M] \rbrace$ is \emph{consistent} with a priority ordering $\PREC$ if, for all pairs of agents where $i \prec j$, we can never improve the arrival time of $a_i$ at $t_i$ by removing $a_j$ from the set of agents.
%A fixed priority ordering $\PREC$ can thus result in a solution that is consistent with it.
\end{defn}

\begin{defn}
A MAPF instance is \textbf{P-solvable} iff there exists
%a solution $L = \lbrace \pi(a_1), \ldots, \pi(a_k) \rbrace$ which is
a solution $L = \lbrace \pi_i~|~i \in [M] \rbrace$ that is
\emph{consistent} with some priority ordering $\PREC$.
\end{defn}

%The next two results are concerned with completeness and optimality of
%prioritized planning, in general, for the set of problems which are P-solvable.
For example, there does not exist any solution that is consistent with any fixed priority ordering for the MAPF instance shown in Figure~\ref{fig:incomplete}.
Thus, the MAPF instance shown in Figure~\ref{fig:incomplete} is not P-solvable.
The next result says that there exist P-solvable MAPF instances with
solutions that are consistent with only one total priority ordering.
%Recall that existing works prioritize agents at random or using informal heuristics.

%Well-formed instances arise naturally in the context of warehouse automation~\cite{CapVK15,MaAAMAS17}
%and set of all such problems are a strict subset of P-solvable.
\begin{thm}
Prioritized planning with any given priority ordering $\PREC$ is incomplete for the class of P-solvable MAPF instance.
\end{thm}
\begin{pf}
Figure~\ref{fig:p_solvable} shows a P-solvable MAPF instance that has only one optimal solution, namely
{\scriptsize
\begin{align*}
    &\pi_1 = \langle~(1, 2), (2, 2), (3, 2), (4,2), (5, 2)~\rangle \\
    &\pi_2 = \langle~(3,2), (3, 1), (3, 2), (2, 2)~\rangle.
\end{align*}}
This solution is only consistent with the
total priority ordering $\lbrace 1 \prec 2 \rbrace$. Prioritized planning for any other priority orderings does not result in a solution.
\end{pf}

The next result focuses on \emph{well-formed} instance, a class
of practical and P-solvable MAPF problems that is important for warehouse
logistics~\cite{CapVK15,MaAAMAS17}.
Figure~\ref{fig:wellformed} shows an example.
The distinguishing feature of this class is that each agent can wait
indefinitely at its start vertex and target vertex without blocking any other agent.

\begin{thm}\label{thm:complete}
Prioritized planning with any given total priority ordering $\PREC$ is complete
for the class of \emph{well-formed} MAPF instances.
\end{thm}
\begin{pf}
Given a well-formed MAPF instance and any total priority ordering $\PREC$,
a consistent solution can always be computed as follows:
every agent waits at its start vertex until all higher priority agents arrive at
their target vertices. Then, the agent follows an individually optimal path from its start vertex to its target vertex.
\end{pf}

\begin{figure}[t!]
  \begin{minipage}[t!]{0.49\columnwidth}
      \includegraphics[width=\columnwidth]{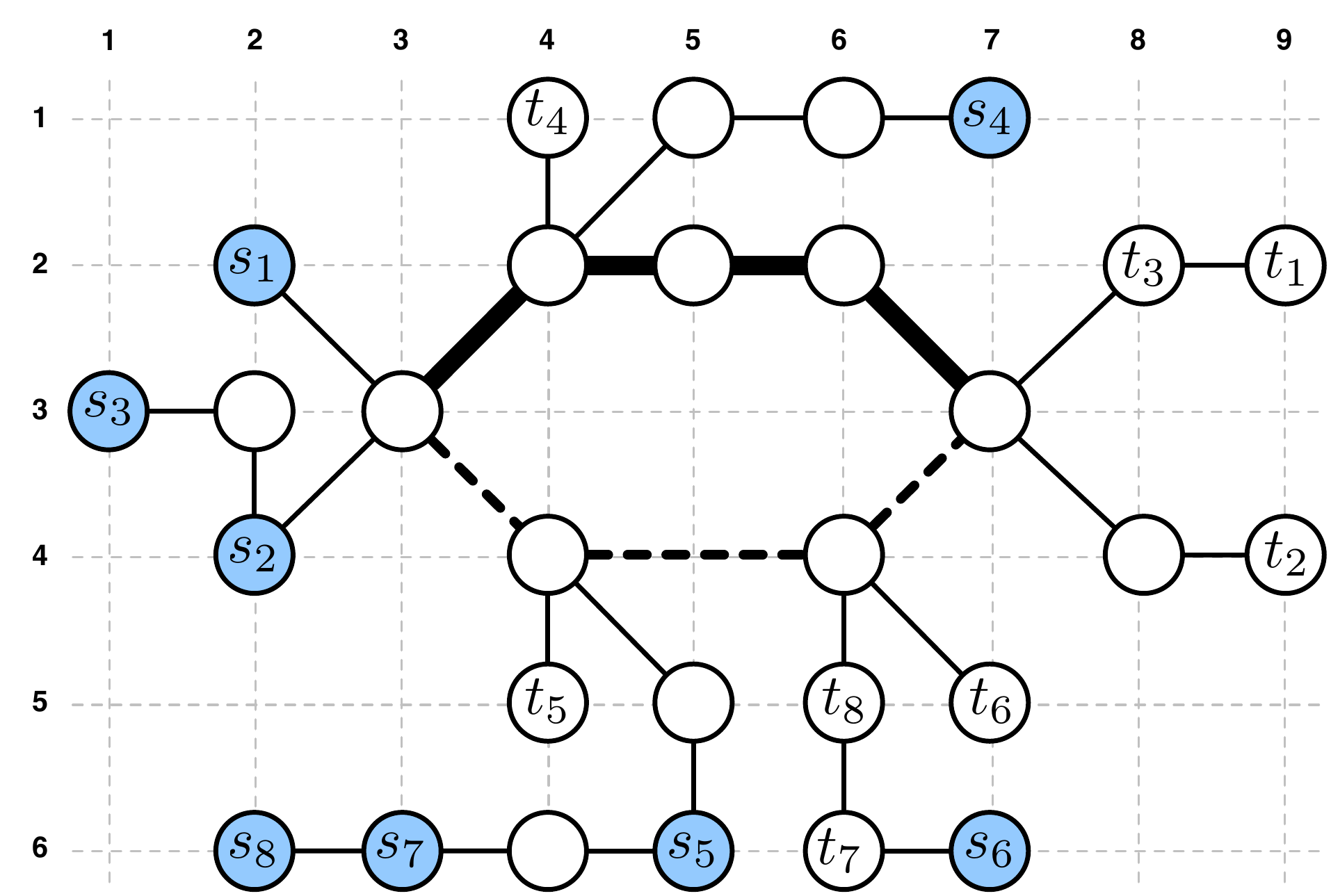}
      \caption{The only flowtime optimal solution for this instance is
          inconsistent with any fixed priority ordering $\PREC$.}
          %The solution requires $a_1 \PREC a_2$ at
          %$(3, 3)$ (timestep 1) and $a_2 \PREC a_1$ at $(7, 3)$ (timestep 5).}
     \label{fig:swaps_soc}
  \end{minipage}
  %\hspace{1cm}
  \hfill
  \begin{minipage}[t!]{0.49\columnwidth}
  \includegraphics[width=\columnwidth]{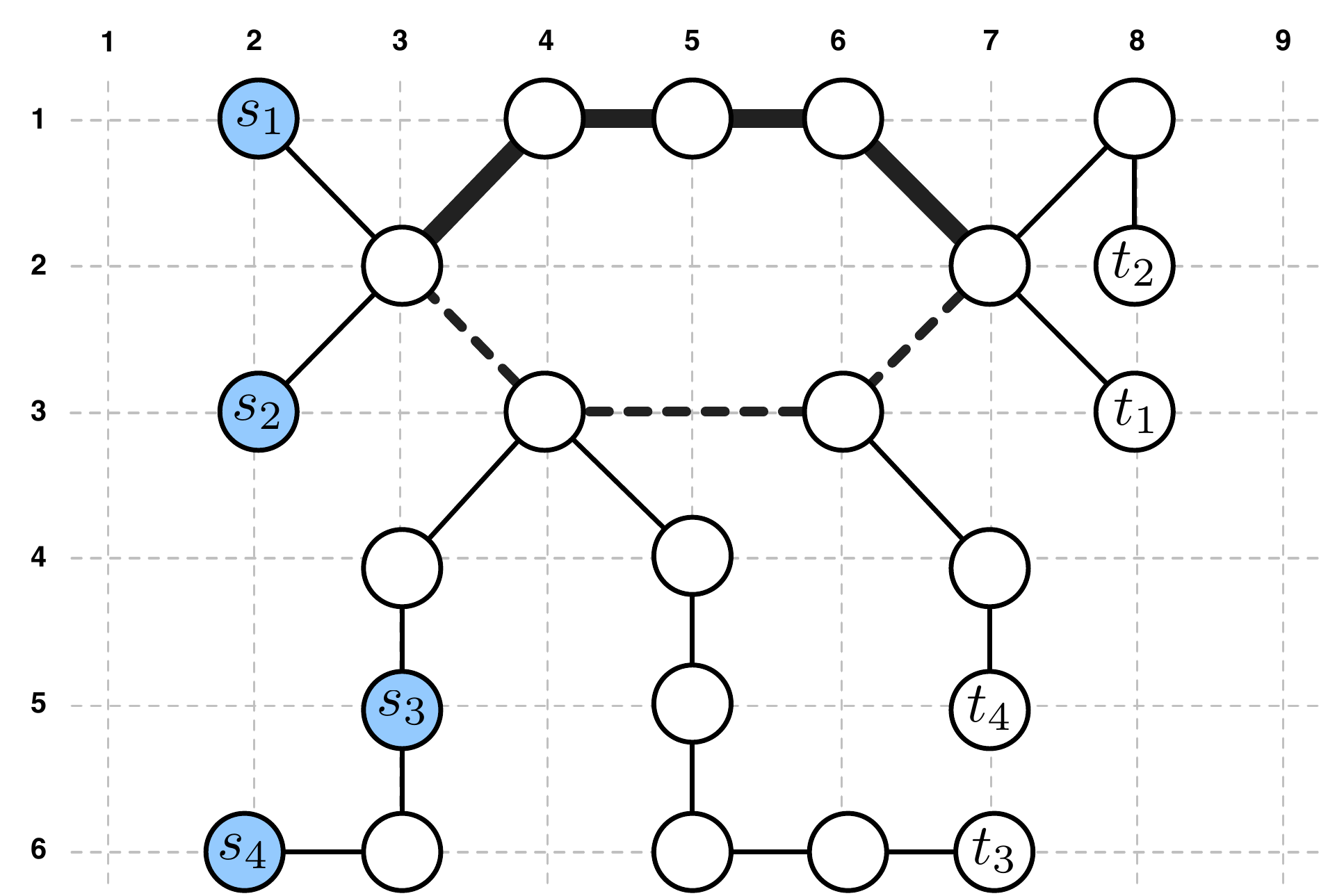}
  \caption{The only makespan optimal solution for this instance is
      inconsistent with any fixed priority ordering $\PREC$.}
 \label{fig:swaps_makespan}
  \end{minipage}
\end{figure}

The next result says that, for some P-solvable MAPF instances (including well-formed ones), prioritized planning can produce suboptimal solutions for all given total priority orderings.
%when using sum-of-costs or makespan as objectives.

\begin{thm}
Prioritized planning
%and prioritized planning using the makespan objective are both suboptimal in
is suboptimal for the flowtime objective in general for the class of P-solvable MAPF instances.
\label{thm:suboptimality}
\end{thm}
\begin{pf}
(Sketch)
The counter-example shown in Figure~\ref{fig:swaps_soc} admits an inconsistent optimal solution $L^{*}$ with flowtime $=48$.
We claim that there exists no priority ordering $\PREC$ which can produce
a consistent prioritized solution with flowtime $<49$.
%Suppose there exists an algorithm $Q$ which, given a
%P-solvable instance $I$, always returns a consistent and cost-optimal
%%prioritized solution $L^{*}_Q = \lbrace \pi(a_1), \ldots, \pi(a_k) \rbrace$
%prioritized solution $L^{*}_Q$ %= \lbrace \pi(a_i)~|~a_i \in [M] \rbrace$
%after considering all possible choices of $\PREC$.
%%$\PREC$ for the set of agents $[M]$.
%We show that there exist some problems $I$ which admit a different and
%inconsistent solution $L^{*}$ which is strictly better than $L^{*}_Q$ under
%sum-of-costs (SOC).
%Figure~\ref{fig:swaps_soc} shows such an instance $I$.
%We sketch an inconsistent optimal solution $L^{*}$ with SOC = 48:
We sketch $L^{*}$ as follows:
\begin{itemize}[nosep]
    \item $1~\prec~2$ at location $(3, 3)$ at time 1.
    \item $a_1$ takes the bold path and arrives at $(6, 2)$ at time 4.
    \item $a_2$ takes the dashed path and arrives at $(6, 4)$ at time 4.
    \item $a_3$ takes the bold path and arrives at $(6, 2)$ at time 6.
    \item $2~\prec~1$ at location $(7, 3)$ at time 5 ($L^{*}$ inconsistent)
    \item $a_2$ reaches its target vertex at time 7; and $a_1$ and $a_3$ reach their target vertices at time 8.
    \item All other agents follow their individually optimal paths.
\end{itemize}
%$L^{*}$ is inconsistent because $a_1$ and $a_2$ swap priorities at
%timestep 5.
%We claim that any optimal prioritized solution $L^{*}_Q$ has SOC $>$ 48.
The proof is by enumeration but the result only depends on the
choice of paths for agents $a_1$, $a_2$ and $a_3$.
All other agents follow their individually optimal paths, waiting or not, depending on their interactions with agents $a_1$, $a_2$ and $a_3$.
For example, if agent $a_1$ chooses the dashed (versus) path it avoids another
crossing with agent $a_2$ and speeds up its arrival at $(7, 3)$.
However, this requires introducing at least two delays due to contention
with agents $a_5$ and $a_6$.
Meanwhile, agent $a_2$ attempts to follow agent $a_1$ but this introduces new
delays from agents $a_5$, $a_6$, $a_7$ and $a_8$.
Alternatively, agent $a_2$ can switch to the bold path, which is slower and introduces a delay due to contention with agent $a_4$.
Both choices lead to solutions with flowtime $>48$.
\end{pf}

\begin{cor}
Theorem~\ref{thm:suboptimality} holds for the makespan objective.
\end{cor}
\begin{pf}
(Sketch) The counter-example shown in Figure~\ref{fig:swaps_makespan} admits
an inconsistent optimal solution and makespan$=7$.
The solution requires $1~\prec~2$ at $(3, 2)$ and $2~\prec~1$ at
$(7, 2)$.
There exists no priority ordering $\PREC$ which can produce
a consistent prioritized solution with makespan$<8$.
The proof is similar to the one of Theorem~\ref{thm:suboptimality}.
\end{pf}

We now focus on a class of MAPF instances for which there exist an
optimal prioritized solution that is also optimal in general.
We refer to such problems as \emph{OP-solvable}.

\begin{defn}
A MAPF instance is \textbf{OP-solvable} iff: (1) it
admits a solution $L^{*}$ that is consistent with some
priority ordering $\PREC^*$ and; (2) $L^{*}$ is optimal among all solutions, whether consistent or not.
\end{defn}

The next result says that prioritized planning may not find an
optimal solution or indeed any solution, even for problems which are
OP-solvable.

\begin{thm}
Prioritized planning with any given fixed total order $\PREC$ is
incomplete in general for the class of problems that are \emph{OP-solvable}.
\end{thm}
\begin{pf}
The counter-example shown in Figure~\ref{fig:op_solvable} admits a consistent
optimal solution (for both the flowtime and makespan objectives) for only the fixed total order
$\lbrace 1 \prec 3 \prec 2 \rbrace$.
The solution is:
{\scriptsize
\begin{align*}
    &\pi_1 = \langle~(2, 1), (3, 1), (4, 2), (5, 1)~\rangle \\
    &\pi_3 = \langle~(2, 2), (2, 2), (3, 1), (4, 2)~\rangle\\
    &\pi_2 = \langle~(4, 1), (4, 1), (4, 1), (3, 1)~\rangle.
\end{align*}}
However, the highest priority agent has another individually optimal path available which involves moving to location $(4, 1)$ at time 2, which results in a deadlock for agent $a_2$.
\end{pf}

%:\begin{defn}
%:A solution $L^{*}_{\PREC}$ to a MAPF problem is \emph{$\PREC$-optimal}
%:if its cost is minimum among all feasible solutions that are consistent
%:with the fixed priority order $\PREC$.
%:A solution $L^{*}_{\PREC}$ to a MAPF problem is \emph{$\PREC$-optimal}
%:When $L^{*}_{\PREC}$ is also minimum cost among all feasible solutions,
%:whether consistent or not, we say the problem is \emph{OP-solvable}.
%:\end{defn}

To summarize: (1) Some MAPF instances that are solvable are not solvable with prioritized planning. (2) Some MAPF instances that are solvable with prioritized planning are only solvable with prioritized planning for a single total priority ordering. (3) Some MAPF instances that are solvable with prioritized planning are not optimally solvable with prioritized planning for any total priority ordering. (4) Even worse, some MAPF instances that are optimally solvable with prioritized planning require prioritized planning not only to use the correct total priority ordering but also break ties correctly when planning paths for the agents, which---if done incorrectly---can prevent prioritized planning from finding any solution.

\section{Conflict-Based Search with Priorities}\label{sec:cbswp}

(Standard) Conflict-Based Search (CBS) is a two-level algorithm that minimizes the flowtime. Conflict-Based Search with Priorities (CBSw/P) is an adaptation of CBS to prioritized planning. Like CBS, CBSw/P performs a best-first search on the high level to resolve collisions among the agents and thus builds a constraint tree (CT). Each CT node $N$ contains a set of constraints $N.\emph{constraints}$, a plan $N.\emph{plan}$ (paths for all agents) that obeys these constraints, and a cost $N.\emph{cost}$ equal to the sum of the arrival times of all paths in its plan at their target vertices. CBSw/P always expands the CT node with the smallest cost. Unlike CBS, CBSw/P also stores a priority ordering $\PREC_{N}$ in each CT node $N$ and only generates child CT nodes $N'$ whose priority orderings $\PREC_{N'}$ extend $\PREC_N$ when it expands a parent CT node $N$.\footnote{We show in the pseudocode how to keep track of $\PREC_{N}$ for each CT node $N$. However, we do not use the notation $N.\PREC_{N}$ but treat it as a global variable for ease of readability.}

\begin{defn}
A priority ordering $\PREC_A$ \emph{extends} a priority ordering $\PREC_B$ if
$\forall i,j \in [M]:~ i \prec_B j \implies i \prec_A j$, that is, $\PREC_A$ maintains all priority information of $\PREC_B$.
\end{defn}

\begin{algorithm}[t]
\setstretch{0}
\KwIn{MAPF instance}
    $\emph{Root}.\emph{constraints}\gets \emptyset$\label{line:CBS:root_constraints}\;
    {\color{red}$\PREC_\emph{Root} \gets \emptyset$}\label{line:CBS:root_ordering}\;
    $\emph{Root}.\emph{plan}\gets$ path for each agent found by a low-level search\label{line:CBS:root_plan}\;
    $\emph{Root}.\emph{cost}\gets$ sum of the arrival times in $\emph{Root}.\emph{plan}$\label{line:CBS:root_key}\;
    $\OPEN \gets \{\emph{Root}\}$\;
    \While{$\OPEN \neq \emptyset$\label{line:CBS:while_statement}}
    {
        $N \gets \arg\min_{N'\in\OPEN}N'.\emph{cost}$\label{line:CBS:choose_node}\;
        $\OPEN \gets \OPEN\setminus\{N\}$\;
        \If{$N.\emph{plan}$ has no collision\label{line:CBS:check_collisions}}
        {
            \Return $N.\emph{plan}$\label{line:CBS:return_solution}
        }
        C $\gets$ a vertex or edge collision $\langle \agent{i},\agent{j},\dots \rangle$ in $N.\emph{plan}$\label{line:CBS:new_collision}\;
        \ForEach{$\agent{i}$ involved in $C$\label{line:CBS:two_child_nodes}}
        {
            %\%\% PJS: Again this has to hold otherwise we should have planned agent $j$ to avoid agent $i$'s path! \;
            \color{red}\If{${i}\nprec_N {j}$ ($\agent{j}$ is the other agent involved in $C$)\label{line:CBS:antisymmetry}}
            {\color{black}
                $N' \gets$ new node\label{line:CBS:new_child}\;            $N'.\emph{plan} \gets N.\emph{plan}$\label{line:CBS:inherit_plan}\;
                $N'.\emph{constraints} \gets N.\emph{constraints} \cup \{\langle\agent{i},\dots\rangle\}$\label{line:CBS:add_constraints}\;
                {\color{red}
                $\PREC_{N'} \gets \PREC_N$\label{line:CBS:inherit_ordering}\;
                %\If{$j \prec_{N'} o_i$}{ \textbf{break} } \; %% alternate
                %\%\% PJS: The test below should have to succeed since if $j \prec_{N'} i$ then we already planned $i$ respecting $j$ and hence a conflict is impossible \;
                \If{${j}\nprec_{N'}{i}$\label{line:CBS:new_ordering}} {
                    $\PREC_{N'} \gets \PREC_{N'} \cup \{{j}\prec {i}\}$\label{line:CBS:add_ordering}\;
                    %$\PREC_{N'} \gets \PREC_{N'} \cup \{{k}\prec {i}|{k}\prec_{N'}{j}, k\in[M]\}$\label{line:CBS:transitive1}\;
                    %$\PREC_{N'} \gets \PREC_{N'} \cup  \{{j}\prec {k}|{i}\prec_{N'}{k}, k\in[M]\}$\label{line:CBS:transitive2}\;
                }
                }
                Update $N'.\emph{plan}$ by invoking a low-level search for $\agent{i}$\label{line:CBS:call_low_level}\;
                \If{a path is returned by the low-level search} {
                $N'.\emph{cost} \gets$  sum of the arrival times in $N'.\emph{plan}$\label{line:CBS:cost}\;
                $\OPEN \gets \OPEN \cup \{N'\}$\label{line:CBS:insert_node}\;
                }
            }
        }
    }
    \Return ``No Solution''\;
\caption{High-Level Search of CBS{\color{red}w/P}}\label{alg:CBS-high}
\end{algorithm}

Algorithm \ref{alg:CBS-high} shows the high-level search of CBSw/P. Red lines are not used in CBS. On the high level, CBSw/P starts with the root CT node, that has an empty set of constraints and an empty priority ordering [Lines \ref{line:CBS:root_constraints}-\ref{line:CBS:root_ordering}]. It performs a low-level search to find an individually optimal path for each agent (without any constraints) independently. The plan of the root CT node thus contains paths for all agents [Line \ref{line:CBS:root_plan}], and its cost is the sum of the arrival times of all paths [Line \ref{line:CBS:root_key}]. When CBSw/P expands a CT node $N$, it checks whether the CT node contains a plan without collisions [Line \ref{line:CBS:check_collisions}]. If this is the case, $N$ is a goal node and CBSw/P terminates successfully [Line \ref{line:CBS:return_solution}]. Otherwise, CBSw/P chooses a collision to resolve [Line \ref{line:CBS:new_collision}] (CBSw/p follows the strategy used in \cite{FelnerICAPS18} to choose a collision) and attempts to generate two candidate child CT nodes $N_1$ and $N_2$ that correspond to the ordered pairs $j\prec i$ and ${i}\prec{j}$, respectively [Line \ref{line:CBS:two_child_nodes}]. It actually generates the child CT node $N_1$ ($N_2$), iff the priority ordering $\PREC_N$ of its parent CT node $N$ does not contain the reversal $i \prec j$ ($j \prec i$) of its corresponding ordered pair $j \prec i$ ($i \prec j$) [Line \ref{line:CBS:antisymmetry}]. Each child CT node inherits the plan, all constraints, and the priority ordering from $N$ [Line \ref{line:CBS:inherit_plan}-\ref{line:CBS:inherit_ordering}].
%\ignore{
If the collision to resolve is a vertex collision $\langle \agent{i}, \agent{j}, v, t\rangle$, CBSw/P adds the vertex constraint $\langle \agent{i}, v, t)$ to $N_1$ (if it is generated) to prohibit agent $\agent{i}$ from occupying $v$ at time $t$ and similarly adds the vertex constraint $\langle \agent{j}, v, t\rangle$ to $N_2$ (if it is generated). If the collision to resolve is an edge collision $\langle \agent{i}, \agent{j}, u, v, t\rangle$, CBSw/P adds the edge constraint $\langle \agent{i}, u, v, t\rangle$ to $N_1$ (if it is generated) to prohibit agent $\agent{i}$ from moving from $u$ to $v$ at time $t$ and similarly adds the edge constraint $\langle \agent{j}, v, u, t\rangle$ to $N_2$ (if it is generated) [Line \ref{line:CBS:add_constraints}].
%}
For each child CT node, say $N_1$, CBSw/P adds its corresponding ordered pair $j\prec i$ to its priority ordering $\PREC_{N_1}$ if the ordered pair is not in $\PREC_{N_1}$ yet [Lines\ref{line:CBS:new_ordering}-\ref{line:CBS:add_ordering}]. CBSw/P uses the same low-level search as CBS (space-time A*) to find an individually optimal path for an agent $\agent{i}$ that respect all constraints in $N_1.\emph{constraints}$ relevant to agent $\agent{i}$. If a new path is found, CBSw/P replaces the old path of agent $\agent{i}$ in $N_1.\emph{plan}$ with the new path returned by the low-level search [Line \ref{line:CBS:call_low_level}], updates the cost of $N_1$ accordingly, and thus inserts $N_1$ into OPEN [Lines \ref{line:CBS:cost}-\ref{line:CBS:insert_node}].

\noindent\textbf{Properties.}
CBSw/P adds a new partially ordered pair to the priority ordering of the child CT node whenever it \emph{splits}, namely generates both child CT nodes of, a parent CT node. Therefore, the number of splitting in any branch of the CT is $O(M^2)$ (the number of all possible ordered pairs). However, the high-level search of CBSw/P is unbounded because the number of all possible collisions between two agents, for each of which a CT node is also expanded, is not finite.

%Note that the depth of the tree explored by CBSw/P is at most $O(n^2)$ for $n$ agents since each level must fix one new partial ordering ordering between 2 agents.  This contrasts with CBS where the depth can be dependent on the size of the map being explored. Note that the total algorithm is still exponential in the worst case, like CBS.

%\noindent\textbf{Bounded Version of CBSw/P}
%When it generates a child CT $N_1$ that corresponds to the ordered pair $j \prec_{N_1} i$, CBSw/P could have specified additional constraints that avoid colliding with the path of agent $a_j$ in $N_1.\emph{plan}$ [Line \ref{line:CBS:add_constraints}]. In this case, CBSw/P expands at most one CT node to resolve all possible collisions between agent $a_{i'}$ with a higher priority and agent $a_{j'}$ with a lower priority after a new priority ordering involving $a_{i'}$ is introduced and a new path for agent $a_{i'}$ is planned. Therefore, the number of expanded nodes for any two partially ordered agents is also bounded. We have tested this version of CBSw/P experimentally. The resulting solution...

\section{Priority-Based Search}\label{sec:pbs}

Priority-Based Search (PBS) is a two-level algorithm for prioritized planning. It performs a depth-first search on the high level to dynamically construct a priority ordering and thus builds a priority tree (PT). Conceptually, when it is faced with a collision, PBS greedily chooses which agent should be given a higher priority. It backtracks efficiently and explores other branches iff there is no solution in the current branch. It thus effectively incrementally constructs a single partial priority ordering until it finds no collisions. PBS shares some similarities with CBSw/P on the high level. Like CBSw/P, PBS splits a PT node and introduces an additional ordered pair only if two agents collide. Unlike CBSw/P, PBS does not store constraints in the PT nodes but maintains the invariant that, if agent $a_i$ has a higher priority than agent $a_j$ ($i \prec_N j$), there are no collisions between them whenever PBS processes a PT node $N$.

%\pjs{We can always treat $\PREC_0$ as input, it can be the empty ordering for normal PBS, and a total ordering otherwise. I think if we fail in the first loop then it must be there is no solution respecting the given partial order!}

\begin{algorithm}[t]
\setstretch{0.4}
\KwIn{MAPF instance, $\PREC_0$($=\emptyset$ by default)}
    $\PREC_\emph{Root} \gets \PREC_0$\label{line:PBS:pre_ordering}\;
    $\emph{Root}.\emph{plan}\gets \emptyset$\label{line:PBS:root_plan}\;
    \ForEach{$i\in[M]$}{
        $\emph{success}\gets\emph{UpdatePlan}(\emph{Root}, a_i)$\label{line:PBS:root_paths}\tcc*{success is always true if $\PREC_0= \emptyset$}%\tcc*{Algorithm \ref{alg:update}}
        \If{$\neg\emph{success}$\label{line:PBS:succ}}{
            \Return ``No Solution''\label{line:PBS:no_solution}\;
        }
    }
    $\emph{Root}.\emph{cost}\gets$ sum of the arrival times in $\emph{Root}.\emph{plan}$\label{line:PBS:root_key}\;
    $\STACK \gets \{\emph{Root}\}$\;
    \While{$\STACK \neq \emptyset$\label{line:PBS:while_statement}}
    {
        $N \gets$ top node in $\STACK$\label{line:PBS:choose_node}\;
        $\STACK \gets \STACK\setminus\{N\}$\;
        \If{$N.\emph{plan}$ has no collision\label{line:PBS:check_collisions}}
        {
            \Return $N.\emph{plan}$\label{line:PBS:return_solution}\;
        }
        C $\gets$ first vertex or edge collision $\langle \agent{i},\agent{j},\dots \rangle$ in $N.\emph{plan}$ \label{line:PBS:new_collision}\;
        \ForEach{$\agent{i}$ involved in $C$\label{line:PBS:two_child_nodes}}
        {
                $N' \gets$ new node\label{line:PBS:new_child}\;            $N'.\emph{plan} \gets N.\emph{plan}$\label{line:PBS:inherit_plan}\;
                $N'.\emph{constraints} \gets N.\emph{constraints} \cup \{\langle\agent{i},\dots\rangle\}$\label{line:PBS:add_constraints}\;
                $\PREC_{N'} \gets \PREC_N \cup \{j\prec i\}$\label{line:PBS:add_ordering}\;
                %Update $N'.\emph{plan}$ by invoking a low-level search for $\agent{i}$\label{line:PBS:call_low_level}\;
                $\emph{success}\gets\emph{UpdatePlan}(N', a_i)$\label{line:PBS:child_paths}\;%\tcc*{Algorithm \ref{alg:update}}
                \If{\emph{success}} {
                    $N'.\emph{cost} \gets$  sum of the arrival times in $N'.\emph{plan}$\label{line:PBS:cost}\;
                }
        }
        Insert new nodes $N'$ into $\STACK$ in non-increasing order of $N'.\emph{cost}$\label{line:PBS:insert_node}\;
    }
    \Return ``No Solution''\;
\Fn{$\textrm{UpdatePlan}(N, a_i)$}{
    $\LIST\gets$ topological sorting on partially ordered set $(\{i\}\cup\{j|i\prec_{N}j\}, \PREC_{N})$\label{line:update:topological}\;
    \ForEach{$j\in\LIST$} {
        \If{$j=i$ or $\exists a_k:k\prec_{N}j$, $a_j$ collides with $a_k$ in $N.\emph{\emph{plan}}$\label{line:update:collide}}{
            Update $N.\emph{plan}$ by invoking a low-level search for $\agent{j}$ that avoids colliding with all agents $a_k$ with higher priorities ($k\prec_{N}j$)\label{line:update:low-level}\;
            \If{no path is returned by the low-level search}{
                \Return \emph{false}\;
            }
        }
    }
    \Return \emph{true}\;
}
\caption{High-Level Search of PBS (with initial priority ordering $\PREC_0$)}\label{alg:PBS-high}
\end{algorithm}

Algorithm \ref{alg:PBS-high} shows the high-level search of PBS. %\pjs{lines 5 and 6 are too close!, can we make the line spacing larger please?}
%Lines in red are used only when a pre-defined priority ordering $\PREC_0$ is given (for example, a total order priority ordering is given in the standard prioritized MAPF algorithm).
We now point out its differences from CBSw/P.
On the high level, PBS starts with the root PT node, that contains an initial priority ordering $\PREC_0$ [Line \ref{line:PBS:pre_ordering}]. For (standard) PBS, this is the empty priority ordering $\emptyset$.
It then calls Function $\emph{UpdatePlan}(N, a_i)$ to find an individually optimal path for each agent $a_i$ [Line \ref{line:PBS:root_paths}] and always succeeds with the empty initial priority ordering [Line \ref{line:PBS:succ}]. When PBS expands a PT node $N$, PBS always generates two child PT nodes $N_1$ and $N_2$ that correspond to the ordered pairs $j\prec i$ and ${i}\prec{j}$, respectively [Line \ref{line:PBS:two_child_nodes}], since agent $a_i$ does not collide with agent $a_j$ if they are comparable with respect to $\PREC_N$ ($i \prec_N j$ or $j \prec_N i$). For each child PT node, say $N_1$, PBS calls Function $\emph{UpdatePlan}(N_1, a_i)$ to find an individually optimal path for agent $a_i$ that avoids collisions with the paths of all higher priority agents [Line \ref{line:PBS:child_paths}]. PBS favors the child PT node with the smallest cost and thus inserts the child PT nodes (if any are generated) into the stack in non-increasing order of their costs [Line \ref{line:PBS:insert_node}].
%\begin{algorithm}[t]
%\setlength\hsize{8.3cm}
%\setstretch{0}
%\Fn{$\textrm{UpdatePlan}(N, a_i)$}{
%    $\LIST\gets$ topological sorting on $(\{i\}\cup\{j|i\prec_{N}j\}, \PREC_{N})$\label{line:update:topological}\;
%    \ForEach{$j\in\LIST$} {
%        \If{$j=i$ or $\exists a_k:k\prec_{N}j$, $a_j$ collides with $a_k$ in $N.\emph{\emph{plan}}$\label{line:update:collide}}{
%            Update $N.\emph{plan}$ by invoking a low-level search for $\agent{j}$ that avoid colliding with all agents $a_k$ with higher priorities ($k\prec_{N}j$)\label{line:update:low-level}\;
%            \If{no path is returned by the low-level search}{
%                \Return \emph{false}\;
%            }
%        }
%    }
%    \Return \emph{true}\;
%}
%\caption{Function $\emph{UpdatePlan}(N, a_i)$}\label{alg:update}
%\end{algorithm}

\noindent\textbf{Function $\bm{\emph{UpdatePlan}(N, a_i)}$.}
%Algorithm \ref{alg:update} shows the pseudocode of Function $\emph{UpdatePlan}(N, a_i)$ that finds a time-minimal path for agent $a_i$ that respects priority ordering $\PREC_N$ and maintains the invariant that the paths of any agent $a_j$ with a lower priority still obey $\PREC_N$.
Function $\emph{UpdatePlan}(N, a_i)$ finds an individually optimal path for agent $a_i$ that avoids collisions with the paths of all higher priority agents and maintains the invariant that the path of any agent $a_j$ with a lower priority still obeys $\PREC_N$ (finds a new path for such an agent $a_j$ if necessary). To do so, PBS performs a topological sort on $i$ and all $j$ with $i\prec_{N}j$ [Line \ref{line:update:topological}]. For each $j$ in the topologically sorted order, PBS then performs a low-level search for agent $\agent{j}$ to compute a new individually optimal path that respects priority ordering $\PREC_N$ if it collides with an agent $\agent{k}$ with a higher priority [Lines \ref{line:update:collide}-\ref{line:update:low-level}].

\noindent\textbf{Low-Level Search.}
On the low level, PBS uses a special low-level search to find an individually optimal path for agent $\agent{j}$ that does not collide with the path of any $\agent{k}$ with a higher priority (i.e., $k\prec_{N}j$). The number of constraints can be infinite since such agent $\agent{k}$ stays in its target vertex $t_k$ forever after its arrival time $T_k$. This is different from the low-level search of CBSw/P where the number of constraints is finite. Therefore, there might be no path that reaches the target vertex $t_j$. PBS thus uses a space-time A* for all times not greater than the maximum of the arrival times of the higher priority agents $a_k$ (i.e., $k\prec_{N}j$) only and switches to standard A* (without the time dimension and wait actions afterward). CBS breaks ties to favor a path that collides with the fewest paths of other agents, which has been empirically shown to make its high-level search more efficient \cite{DBLP:journals/ai/SharonSFS15}. Similarly, PBS breaks ties first to favor a path that collides with the fewest paths of agents $\agent{k'}$ that are not comparable with agent $\agent{j}$ with respect to $\PREC_N$ and second to favor a path that collides with the fewest paths of lower priority agents $\agent{k''}$ (i.e., $j\prec_{N}k''$) to avoid having to replan a path for such agents later as much as possible.

\noindent\textbf{Properties.}
PBS introduces a new ordered pair to the priority ordering of the child PT node whenever it splits a parent PT node. Therefore, the number of splits in any branch of the PT is $O(M^2)$ (the number of all possible ordered pairs). The depth of the PT is also $O(M^2)$ since PBS splits a PT node whenever it expands the PT node. In practice, PBS expands many fewer PT nodes for each MAPF instance where it finds a solution in our experiments.

\noindent\textbf{Other Versions of PBS.}
A non-empty initial priority ordering $\PREC_0$ [Line \ref{line:PBS:pre_ordering}] can be given so that the solution found by PBS must respect $\PREC_0$.
%This requirement arises in many real-world applications. For example, in an automated warehouse, robots carrying shelves are of higher priorities than robots on the way to their assigned shelves because the former have to deliver the shelves but the latter can be reassigned different shelves (before they pick up the shelves) and thus change the paths.
However, there might not exist any solutions that are consistent with $\PREC_0$. PBS might thus terminate in the root PT node [Line \ref{line:PBS:no_solution}]. For example, for the MAPF instance shown in \cref{fig:p_solvable}, PBS with the initial priority ordering $\PREC_0=\{2\prec 1\}$ does not find any solution but standard PBS (the empty initial priority ordering) constructs the priority ordering $\PREC =\{1\prec 2\}$ and finds a solution. A standard prioritized MAPF algorithm is a special case of PBS where the initial priority ordering is a total order.

\section{Experiments}

We compare CBSw/P and PBS to CBS and several PBS variants that simulate standard prioritized MAPF algorithms with different total priority orderings. We consider the following algorithms: CBS-H (labeled \textbf{CBS}) is a state-of-the-art implementation of CBS~\cite{FelnerICAPS18}; \textbf{FIX} is a PBS variant with a total priority ordering specified by the order of the agents in the MAPF instances, which simulates CA*~\cite{WHCA}; \textbf{LH} is a PBS variant with a fixed total priority ordering where the priority of an agent is higher the longer its individually optimal path from its start vertex to its target is, which simulates the heuristic used in in~\cite{van2005prioritized}; \textbf{SH} a PBS variant with a fixed total priority ordering where the priority of an agent is higher the shorter its individually optimal path from its start vertex to its target vertex is, which is the opposite of the heuristic of LH; and \textbf{RND} is a PBS variant that runs PBS ten times with different randomly generated total priority orderings and picks the solution with the smallest cost, which simulates the randomized strategy in~\cite{bennewitz2002finding}.
All experiments were run on a 2.50 GHz Intel Core i5-2450M laptop with 6 GB RAM and a runtime limit of one minute for each MAPF instance for each algorithm except that RND was given a runtime limit of one minute for each of its ten runs. We repeated each experiment 50 times for each number of agents using different randomly generated start and target vertices for the agents, and report the mean values.

\begin{figure}[t!]
\centering
\begin{subfigure}[b]{.5\columnwidth}
\centering
  \includegraphics[width=\columnwidth]{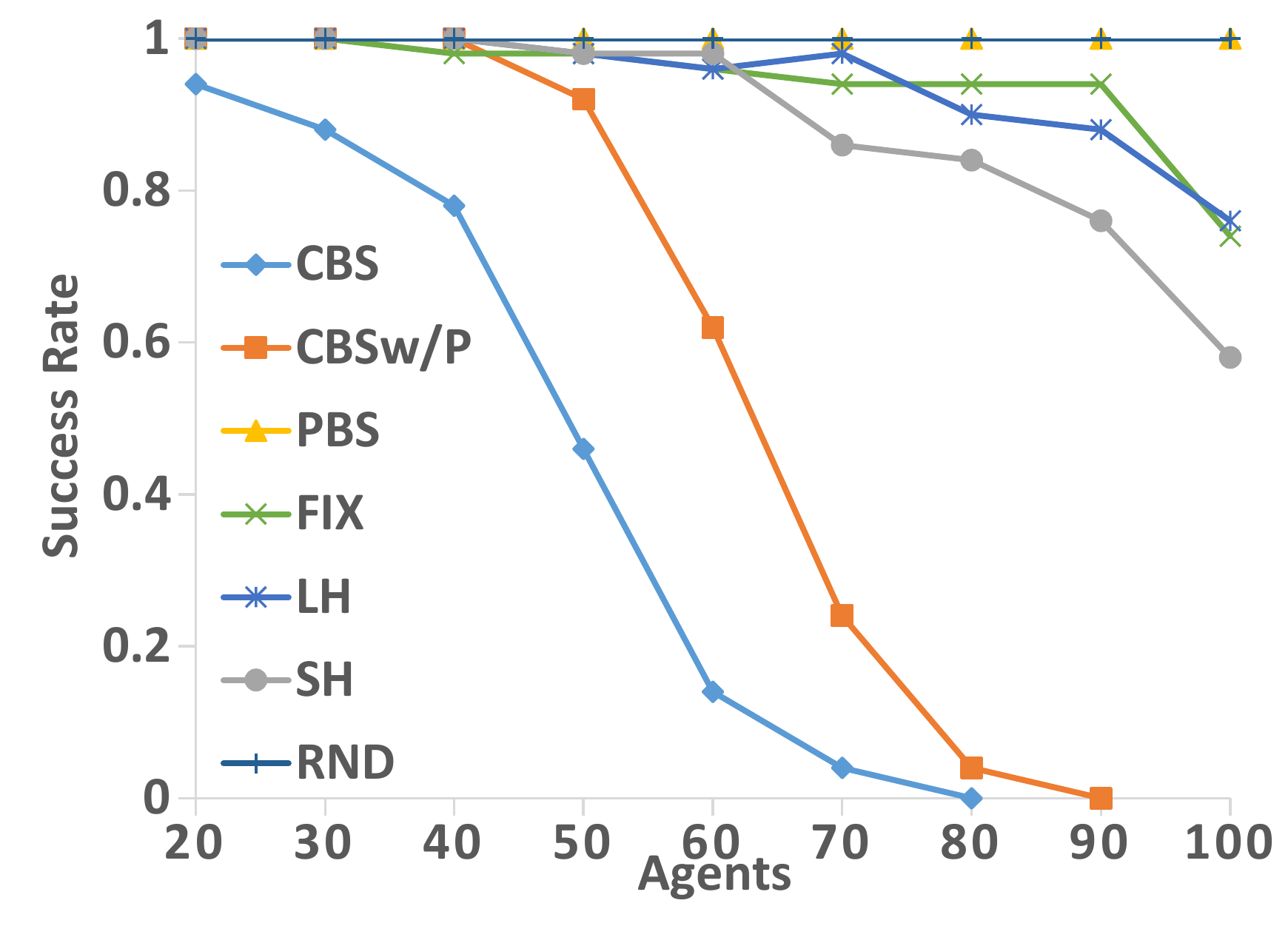}
  \caption{Success rates for 0\% obstacles.}\label{fig:succ_20}
\end{subfigure}%
\begin{subfigure}[b]{.5\columnwidth}
\centering
  \includegraphics[width=\columnwidth]{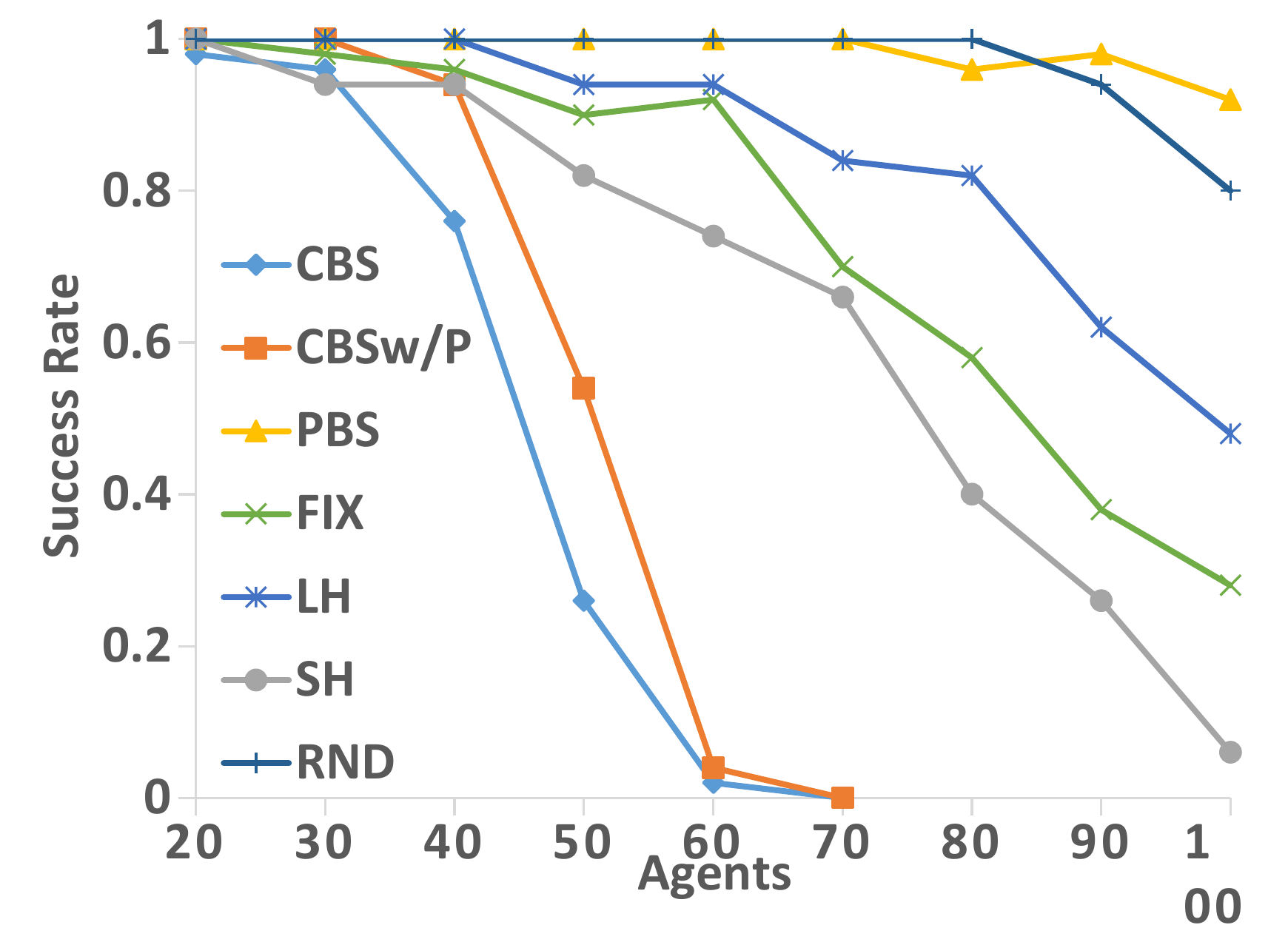}
  \caption{Success rates for 10\% obstacles.}\label{fig:succ_20_obs}
\end{subfigure}\\
\begin{subfigure}[b]{.5\columnwidth}
  \includegraphics[width=\columnwidth]{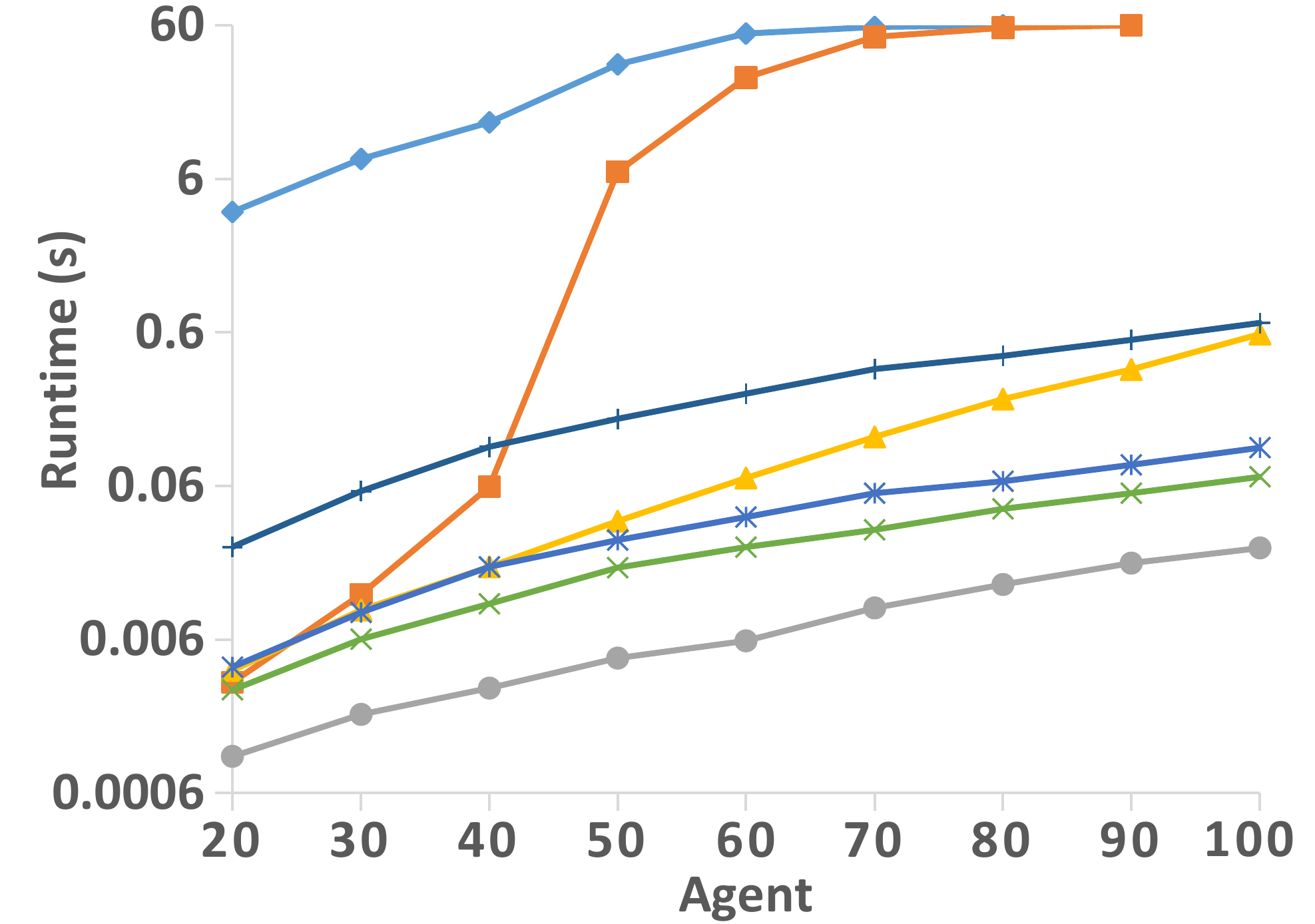}
  \caption{Runtimes (logarithmic) for 0\% obstacles.}\label{fig:time_20}
\end{subfigure}%
\begin{subfigure}[b]{.5\columnwidth}
  \includegraphics[width=\columnwidth]{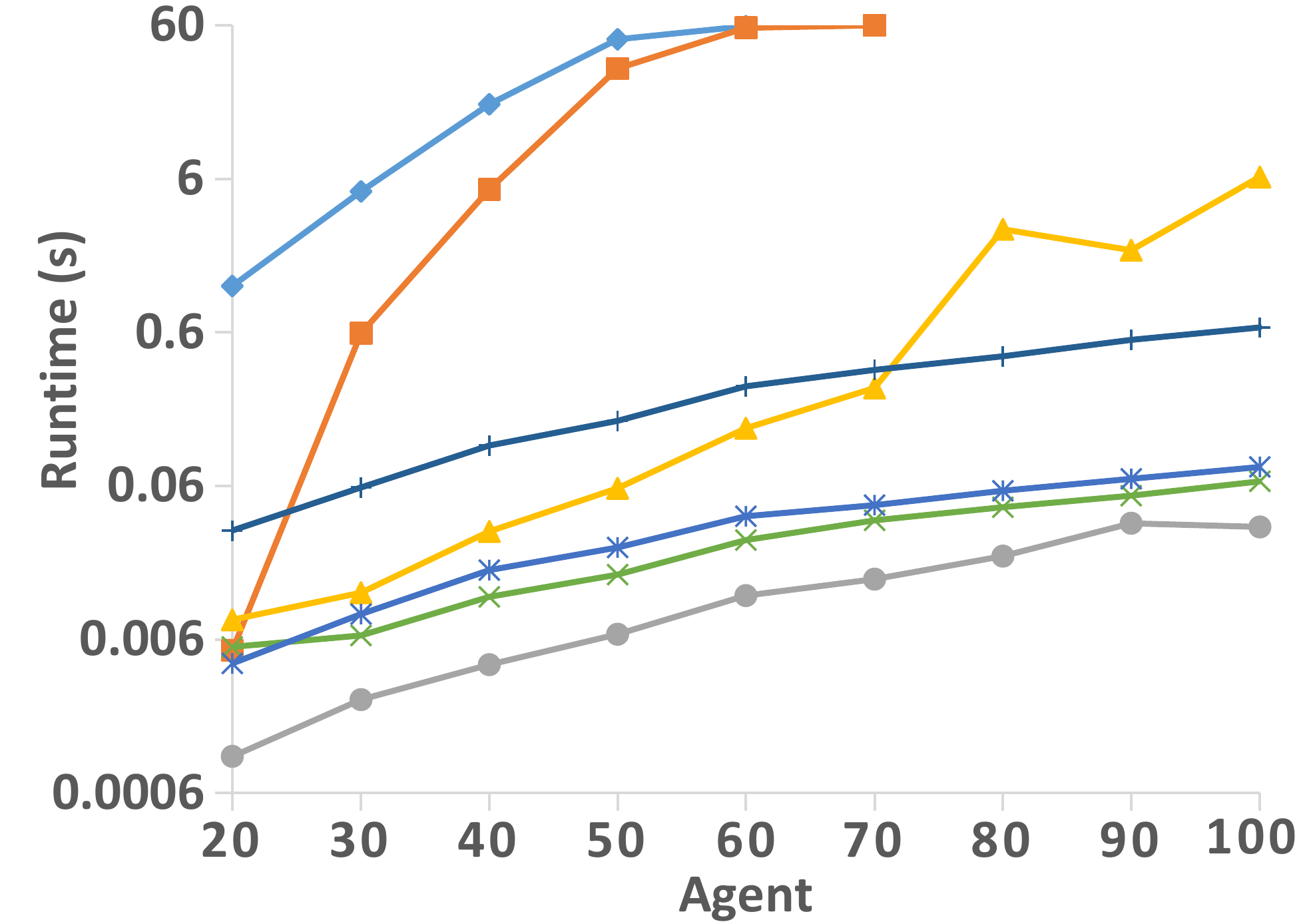}
  \caption{Runtimes (logarithmic) for 10\% obstacles.}\label{fig:time_20_obs}
\end{subfigure}\\
\begin{subfigure}[b]{.5\columnwidth}
  \includegraphics[width=\columnwidth]{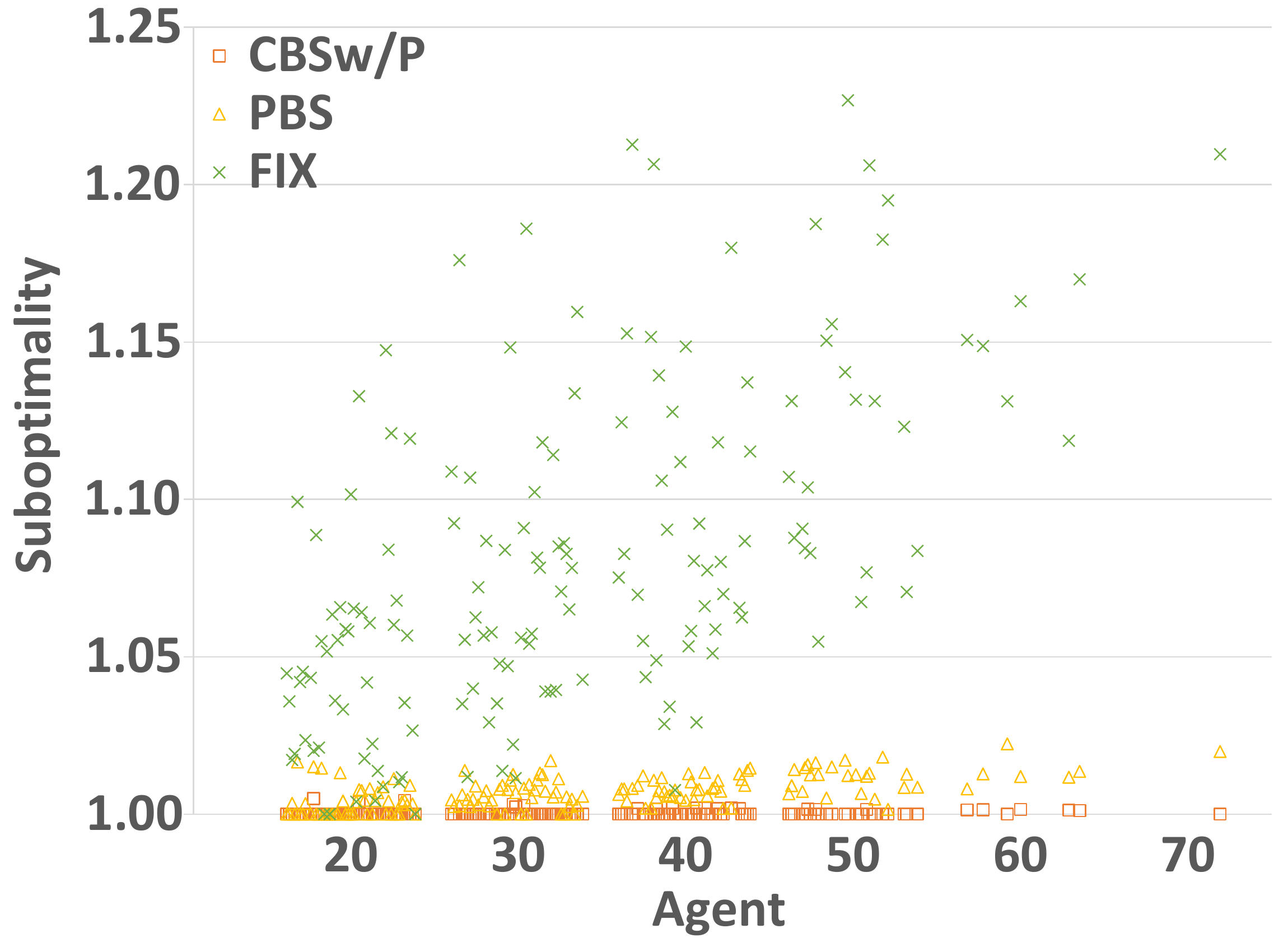}
  \caption{Suboptimality ratios for 0\% obstacles.}\label{fig:sub_20}
\end{subfigure}%
\begin{subfigure}[b]{.5\columnwidth}
  \includegraphics[width=\columnwidth]{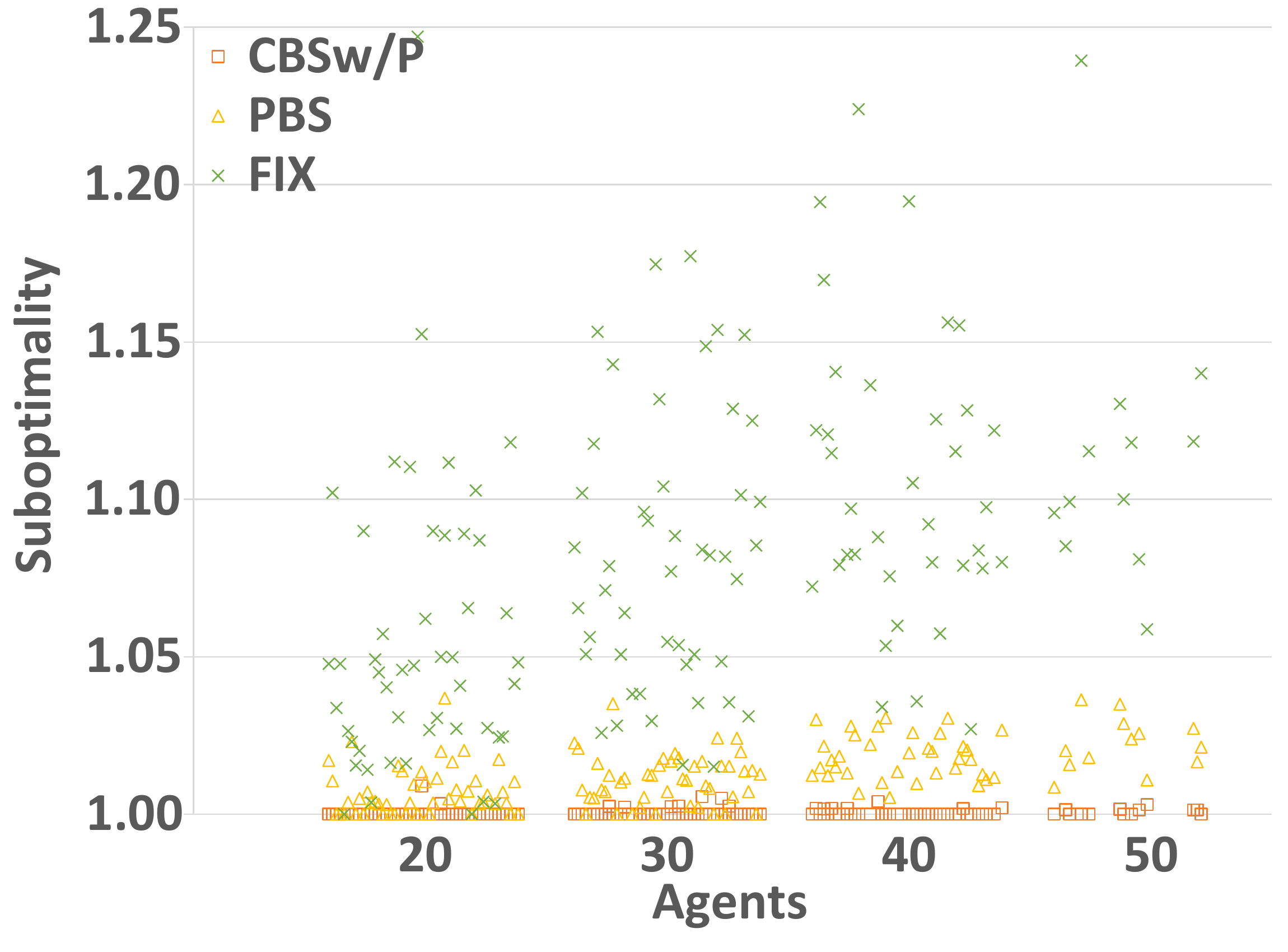}
  \caption{Suboptimality ratios for 10\% obstacles.}\label{fig:sub_20_obs}
\end{subfigure}\\
\begin{subfigure}[b]{\columnwidth}
\setlength{\tabcolsep}{.75pt}
  %\bf\tiny
  \Large
  \resizebox{\columnwidth}{!}{
\begin{tabular}{@{\hspace{-\tabcolsep}}c@{\hspace{-\tabcolsep}}|c|r|r|r|r|r|r|r|r|r|r|r|r}
\hline
\multicolumn{1}{c|}{obs} & \multicolumn{1}{c|}{$M$} & sol & \multicolumn{4}{c|}{\begin{tabular}[c]{@{\hspace{-\tabcolsep}}c@{\hspace{-\tabcolsep}}}low-level node expansions\\(\bm{$\times 1000$})\end{tabular}}                                                                                            & \multicolumn{3}{c|}{\begin{tabular}[c]{@{\hspace{-\tabcolsep}}c@{\hspace{-\tabcolsep}}}high-level node\\ expansions\end{tabular}}                                                           & \multicolumn{4}{c}{flowtime}                                                                                                \\
\hline
\multicolumn{1}{c|}{}       & \multicolumn{1}{c|}{}       & & \multicolumn{1}{c}{CBS} & \multicolumn{1}{c}{\begin{tabular}[c]{@{}l@{}}CBS\\ w/P\end{tabular}} & \multicolumn{1}{c}{PBS} & \multicolumn{1}{c|}{FIX} & \multicolumn{1}{c}{CBS} & \multicolumn{1}{c}{\begin{tabular}[c]{@{}l@{}}CBS\\ w/P\end{tabular}} & \multicolumn{1}{c|}{PBS} & \multicolumn{1}{c}{CBS} & \multicolumn{1}{c}{\begin{tabular}[c]{@{}l@{}}CBS\\ w/P\end{tabular}} & \multicolumn{1}{c}{PBS} & \multicolumn{1}{c}{FIX} \\
\hline
\multirow{6}{*}{0\%}  & 20 & 47 & 22.02     & 0.68     & 1.04  & 0.86  & 268.45     & 7.74      & 3.43  & 252.83 & 252.89 & 253.68 & 265.15   \\
                      & 30 & 44 & 369.88    & 2.69     & 2.38  & 1.92  & 4,319.11   & 40.98     & 7.68  & 399.14 & 399.20 & 401.52 & 428.32   \\
                      & 40 & 38 & 353.21    & 13.50    & 3.79  & 2.95  & 5,430.03   & 161.68    & 12.97 & 519.32 & 519.53 & 523.42 & 566.55   \\
                      & 50 & 23 & 778.13    & 99.97    & 6.65  & 5.23  & 10,813.09  & 1,167.74  & 23.74 & 661.57 & 661.70 & 669.00 & 744.57   \\
                      & 60 & 6  & 3,428.58  & 362.16   & 9.62  & 6.50  & 60,338.67  & 5,805.00  & 36.50 & 741.00 & 741.83 & 750.83 & 850.33   \\
                      & 70 & 1  & 13,337.57 & 3,627.07 & 19.46 & 12.04 & 184,554.00 & 58,408.00 & 46.00 & 854.00 & 854.00 & 871.00 & 1,033.00 \\
                      \hline
\multirow{4}{*}{10\%} & 20 & 49 & 1.30      & 0.86     & 1.37  & 1.02  & 16.16      & 9.14      & 4.69  & 271.02 & 271.08 & 272.90 & 285.57   \\
                      & 30 & 47 & 674.99    & 51.26    & 3.25  & 2.16  & 5,219.55   & 454.15    & 11.38 & 404.40 & 404.60 & 408.81 & 437.47   \\
                      & 40 & 36 & 1,721.40  & 318.47   & 5.96  & 3.22  & 21,688.75  & 4,981.56  & 24.61 & 539.17 & 539.39 & 548.89 & 595.42   \\
                      & 50 & 13 & 5,197.59  & 680.32   & 8.37  & 4.86  & 63,240.23  & 9,742.54  & 37.23 & 675.92 & 676.46 & 691.00 & 762.77    \\
\hline
\end{tabular}
}
\caption{Results on instances solved by all CBS, CBSw/P, PBS, and FIX.\label{tab:2020_top}}
\end{subfigure}\\
\begin{subfigure}[b]{.5\columnwidth}
\centering
  \includegraphics[width=\columnwidth]{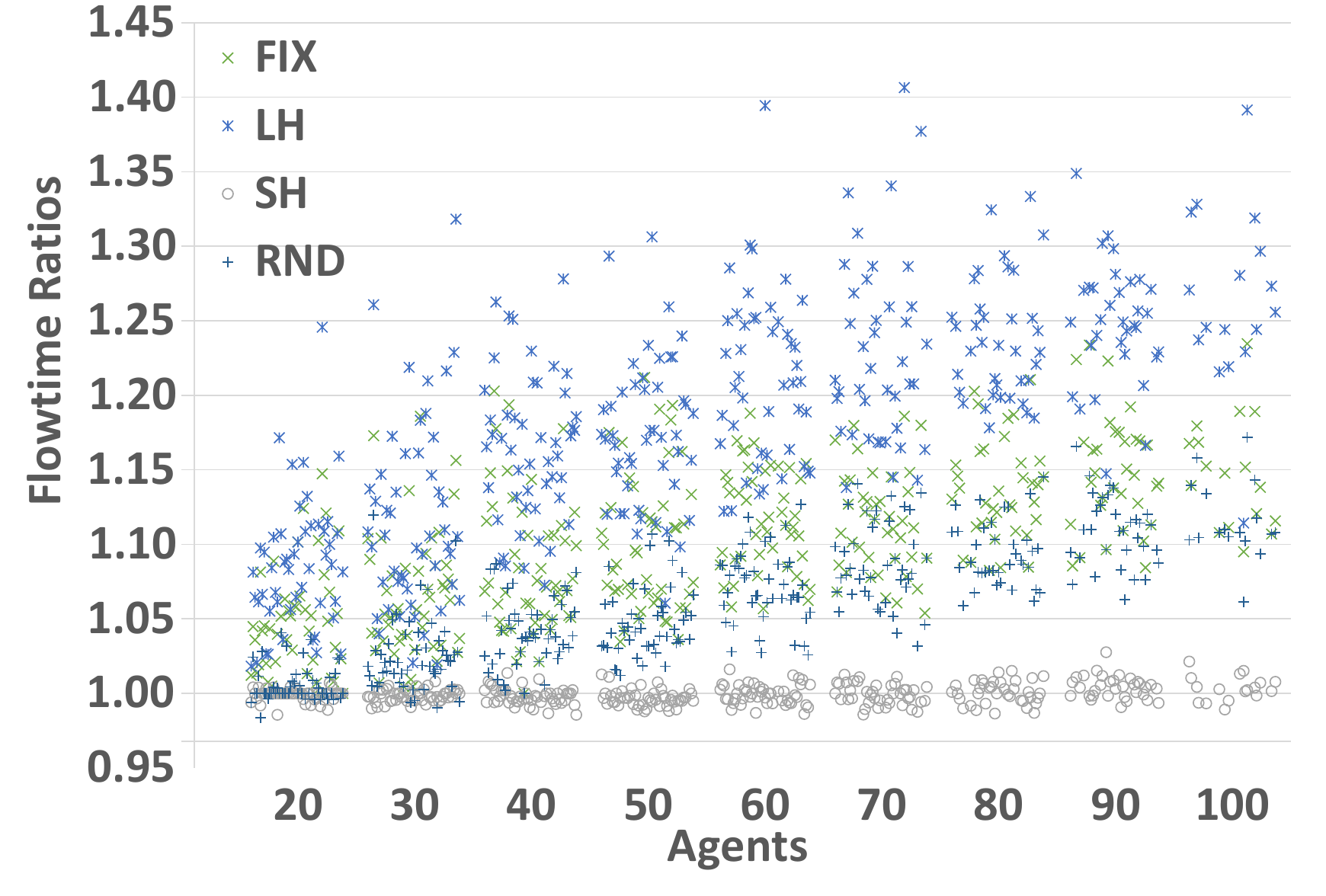}
  \caption{Flowtime ratios against PBS for 0\% obstacles.}\label{fig:rat_20}
\end{subfigure}%
\begin{subfigure}[b]{.5\columnwidth}
\centering
  \includegraphics[width=\columnwidth]{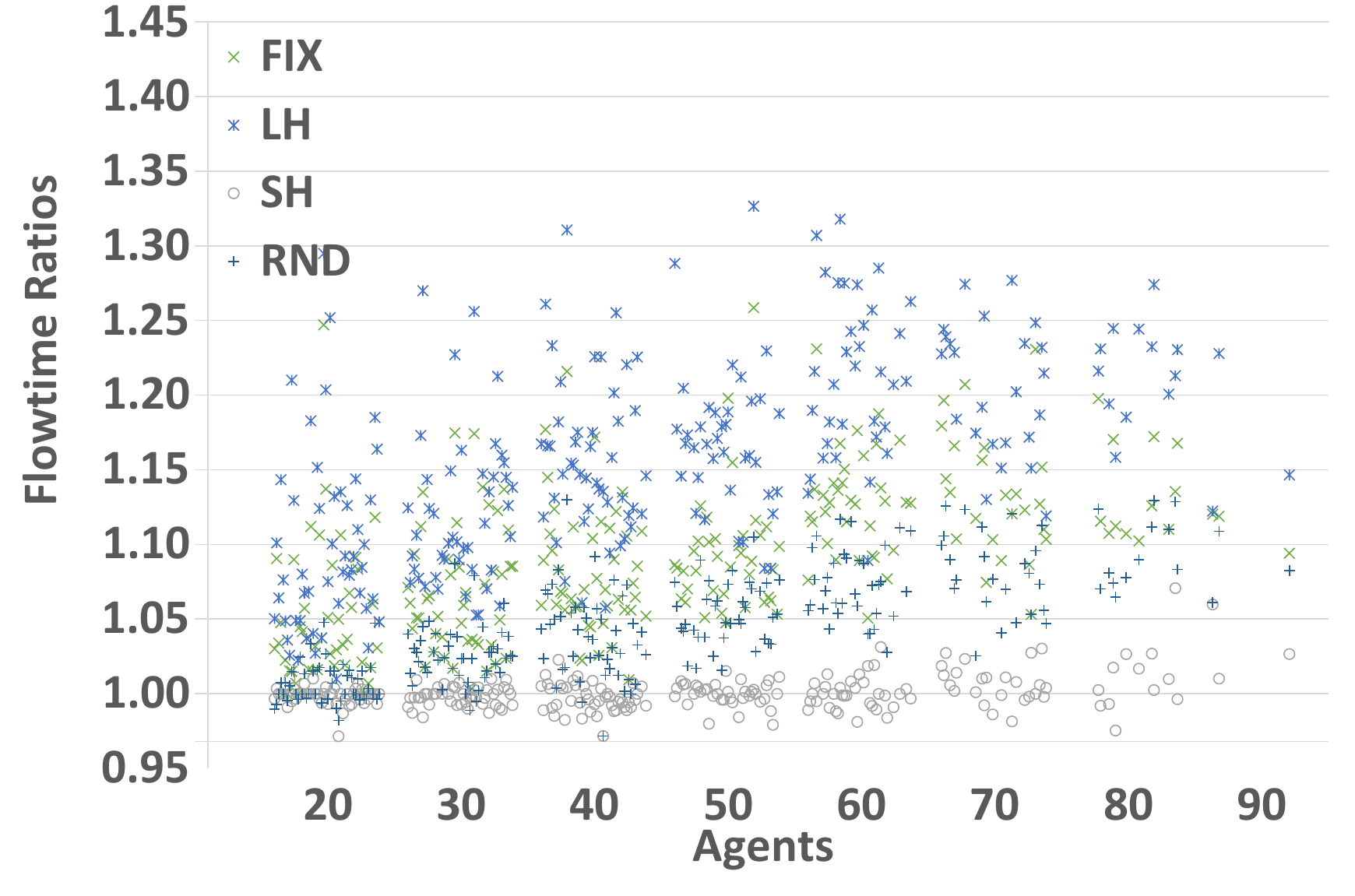}
  \caption{Flowtime ratios against PBS for 10\% obstacles.}\label{fig:rat_20_obs}
\end{subfigure}\\
\begin{subfigure}[b]{\columnwidth}
\setlength{\tabcolsep}{1.5pt}
  %\bf\tiny
  \Large
\centering
\resizebox{\columnwidth}{!}{
\begin{tabular}{@{\hspace{-\tabcolsep}}c@{\hspace{-\tabcolsep}}|r|r|r|r|r|r|r|r|r|r|r|r}
\hline
\multicolumn{1}{c|}{obs} & \multicolumn{1}{c|}{$M$} & \multicolumn{1}{c|}{sol} & \multicolumn{5}{c|}{\begin{tabular}[c]{@{\hspace{-\tabcolsep}}c@{\hspace{-\tabcolsep}}}low-level node expansions\\(\bm{$\times 1000$})\end{tabular}}                                                                                 & \multicolumn{5}{c}{flowtime}                                                                                                      \\
\hline
\multicolumn{1}{c|}{}    & \multicolumn{1}{c|}{}    & \multicolumn{1}{c|}{}    & \multicolumn{1}{c}{PBS} & \multicolumn{1}{c}{FIX} & \multicolumn{1}{c}{LH} & \multicolumn{1}{c}{SH} & \multicolumn{1}{c|}{RND} & \multicolumn{1}{c}{PBS} & \multicolumn{1}{c}{FIX} & \multicolumn{1}{c}{LH} & \multicolumn{1}{c}{SH} & \multicolumn{1}{c}{RND} \\
\hline
\multirow{9}{*}{0\%}  & 20  & 50 & 1.03  & 0.88  & 1.29  & 0.32  & 8.35   & 255.84   & 267.30   & 277.50   & 255.56   & 257.38   \\
                      & 30  & 50 & 2.34  & 1.87  & 2.99  & 0.54  & 18.10  & 401.26   & 427.08   & 447.60   & 400.58   & 411.34   \\
                      & 40  & 49 & 3.82  & 3.08  & 5.44  & 0.76  & 32.81  & 526.96   & 571.41   & 612.67   & 525.84   & 548.88   \\
                      & 50  & 48 & 6.91  & 4.85  & 7.84  & 1.08  & 45.99  & 675.27   & 743.94   & 792.96   & 673.04   & 707.67   \\
                      & 60  & 46 & 11.46 & 6.40  & 10.87 & 1.46  & 63.84  & 798.33   & 891.26   & 965.67   & 797.78   & 855.35   \\
                      & 70  & 41 & 17.62 & 7.57  & 13.99 & 2.04  & 85.94  & 955.41   & 1,064.02 & 1,172.61 & 954.83   & 1,037.95 \\
                      & 80  & 37 & 26.85 & 9.67  & 15.17 & 2.69  & 94.86  & 1,076.65 & 1,222.16 & 1,330.78 & 1,077.24 & 1,177.43 \\
                      & 90  & 33 & 41.66 & 12.62 & 19.12 & 3.80  & 121.57 & 1,258.42 & 1,444.24 & 1,571.06 & 1,263.67 & 1,394.91 \\
                      & 100 & 17 & 59.28 & 14.63 & 24.90 & 5.08  & 156.56 & 1,420.18 & 1,629.65 & 1,793.47 & 1,425.12 & 1,584.65 \\
                      \hline
\multirow{8}{*}{10\%} & 20  & 50 & 1.37  & 1.00  & 1.67  & 0.38  & 10.46  & 273.84   & 286.26   & 299.90   & 273.44   & 275.66   \\
                      & 30  & 46 & 3.16  & 2.11  & 3.04  & 0.67  & 19.82  & 410.26   & 438.65   & 458.26   & 409.57   & 421.26   \\
                      & 40  & 47 & 6.67  & 3.35  & 5.47  & 1.09  & 34.56  & 552.26   & 598.30   & 638.91   & 550.74   & 573.49   \\
                      & 50  & 38 & 11.24 & 4.60  & 7.06  & 1.64  & 45.62  & 703.79   & 769.53   & 822.53   & 703.76   & 741.26   \\
                      & 60  & 34 & 23.19 & 7.75  & 11.29 & 2.72  & 73.93  & 843.44   & 951.12   & 1,022.79 & 843.41   & 905.21   \\
                      & 70  & 24 & 42.07 & 10.16 & 13.31 & 4.29  & 92.90  & 1,012.79 & 1,148.13 & 1,219.33 & 1,019.88 & 1,094.63 \\
                      & 80  & 12 & 60.07 & 11.95 & 16.59 & 6.22  & 111.44 & 1,188.83 & 1,349.50 & 1,448.42 & 1,201.17 & 1,301.75 \\
                      & 90  & 3  & 97.01 & 14.48 & 16.20 & 11.12 & 128.01 & 1,406.00 & 1,562.33 & 1,638.00 & 1,451.67 & 1,524.00   \\
\hline
\end{tabular}
}
\caption{Results on instances solved by all PBS, FIX, LH, SH, and RND.\label{tab:2020_bot}}
\end{subfigure}
\caption{Results on 20$\times$20 grids.\label{fig:2020}}
\end{figure}

\begin{figure}[htbp]
\centering
\begin{subfigure}[b]{.5\columnwidth}
\centering
  \includegraphics[width=\columnwidth]{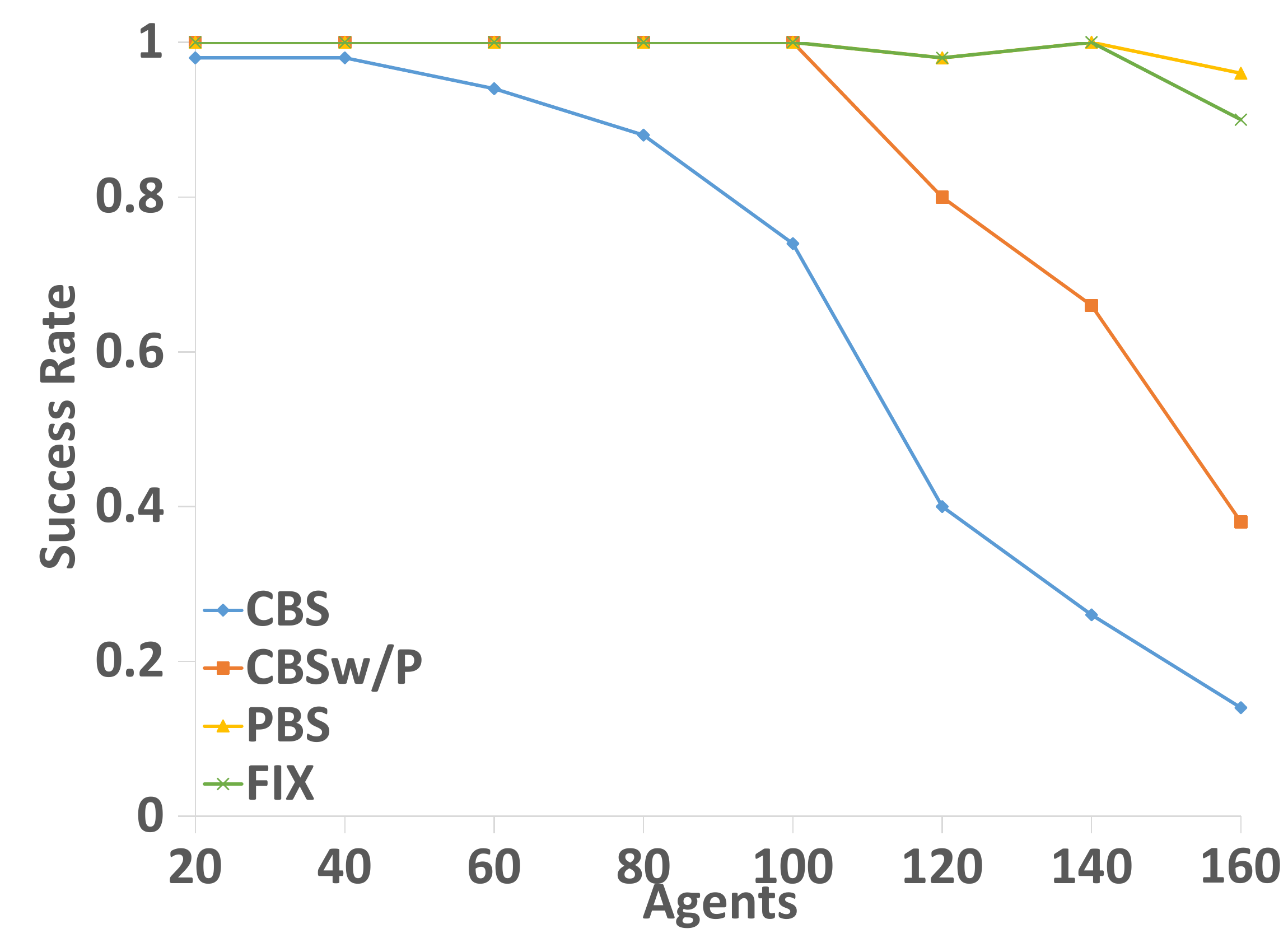}
  \caption{Success rates for brc202d.}
\end{subfigure}%
\begin{subfigure}[b]{.5\columnwidth}
\centering
  \includegraphics[width=\columnwidth]{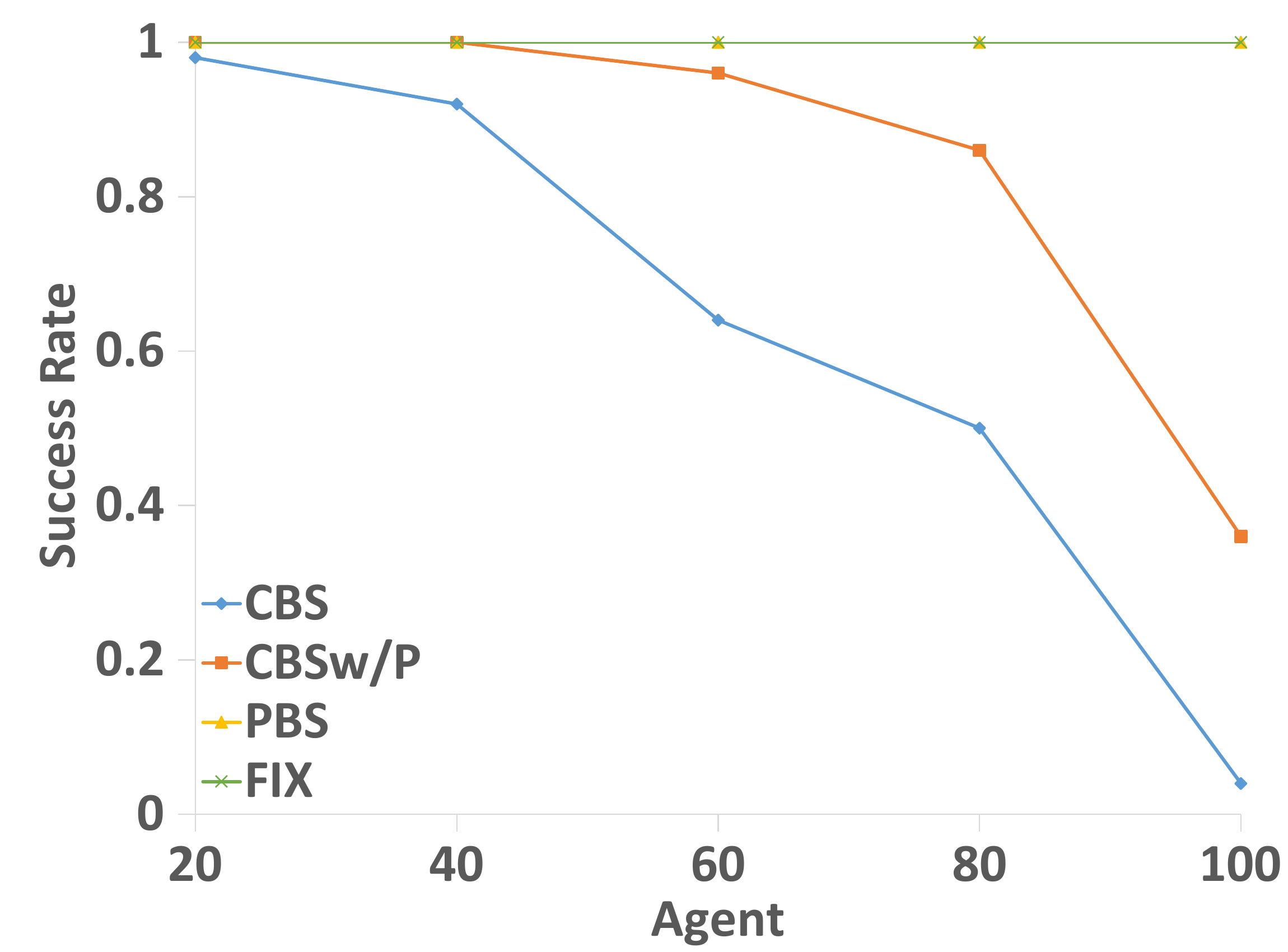}
  \caption{Success rates for lak503d.}
\end{subfigure}\\
\begin{subfigure}[b]{.5\columnwidth}
  \includegraphics[width=\columnwidth]{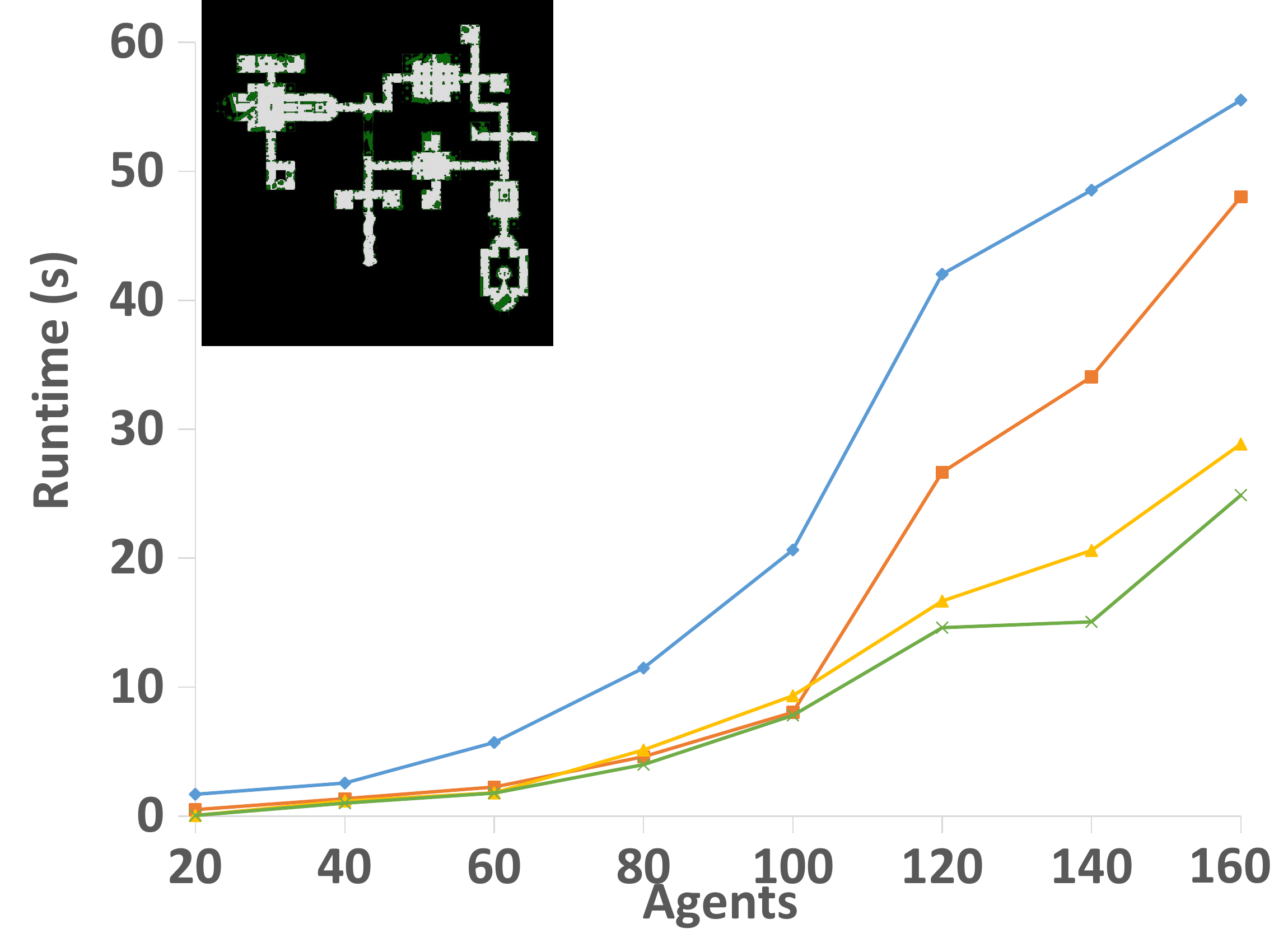}
  \caption{Runtimes for brc202d.}
\end{subfigure}%
\begin{subfigure}[b]{.5\columnwidth}
  \includegraphics[width=\columnwidth]{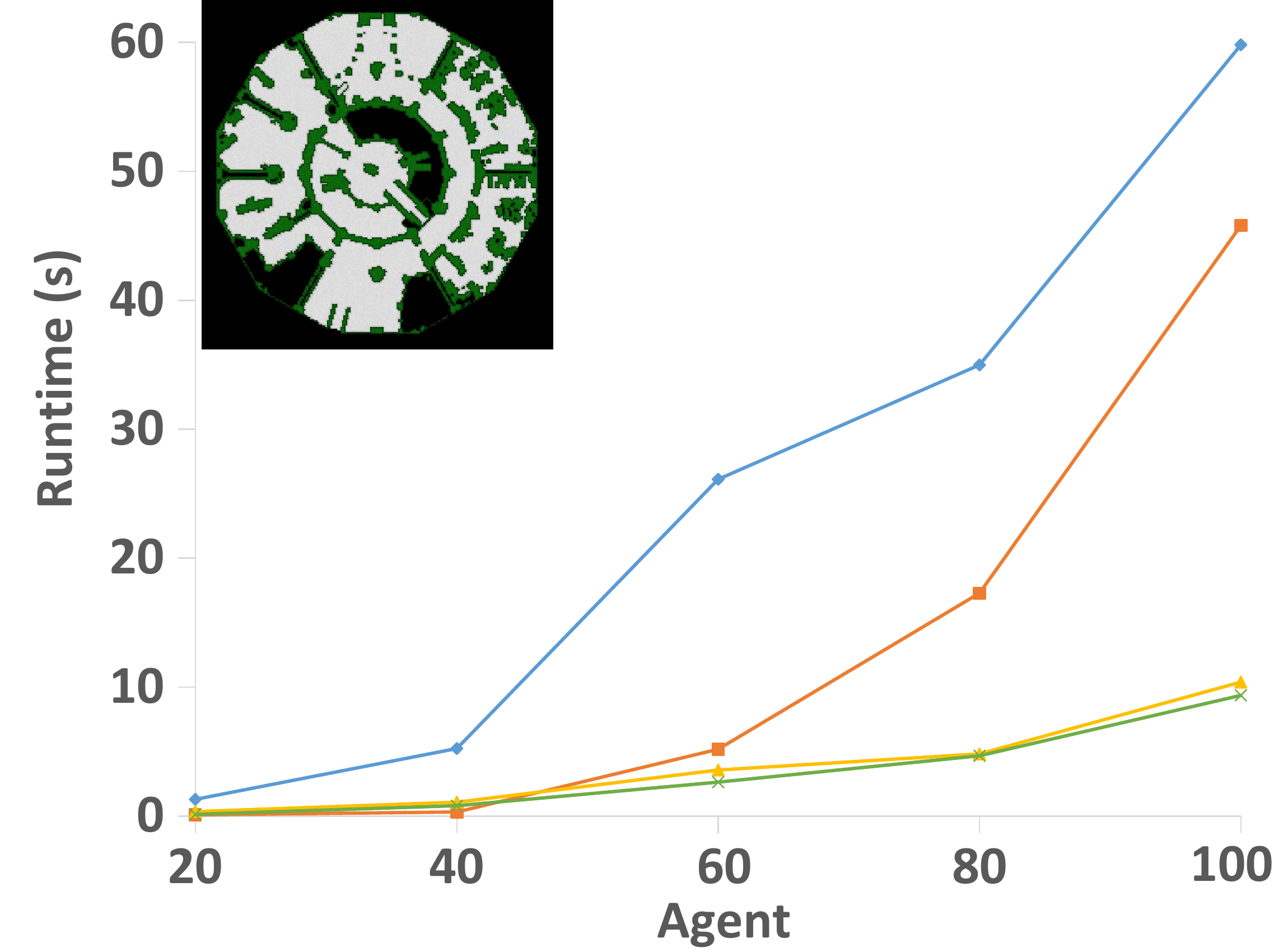}
  \caption{Runtimes for lak503d.}
\end{subfigure}\\
\begin{subfigure}[b]{.5\columnwidth}
  \includegraphics[width=\columnwidth]{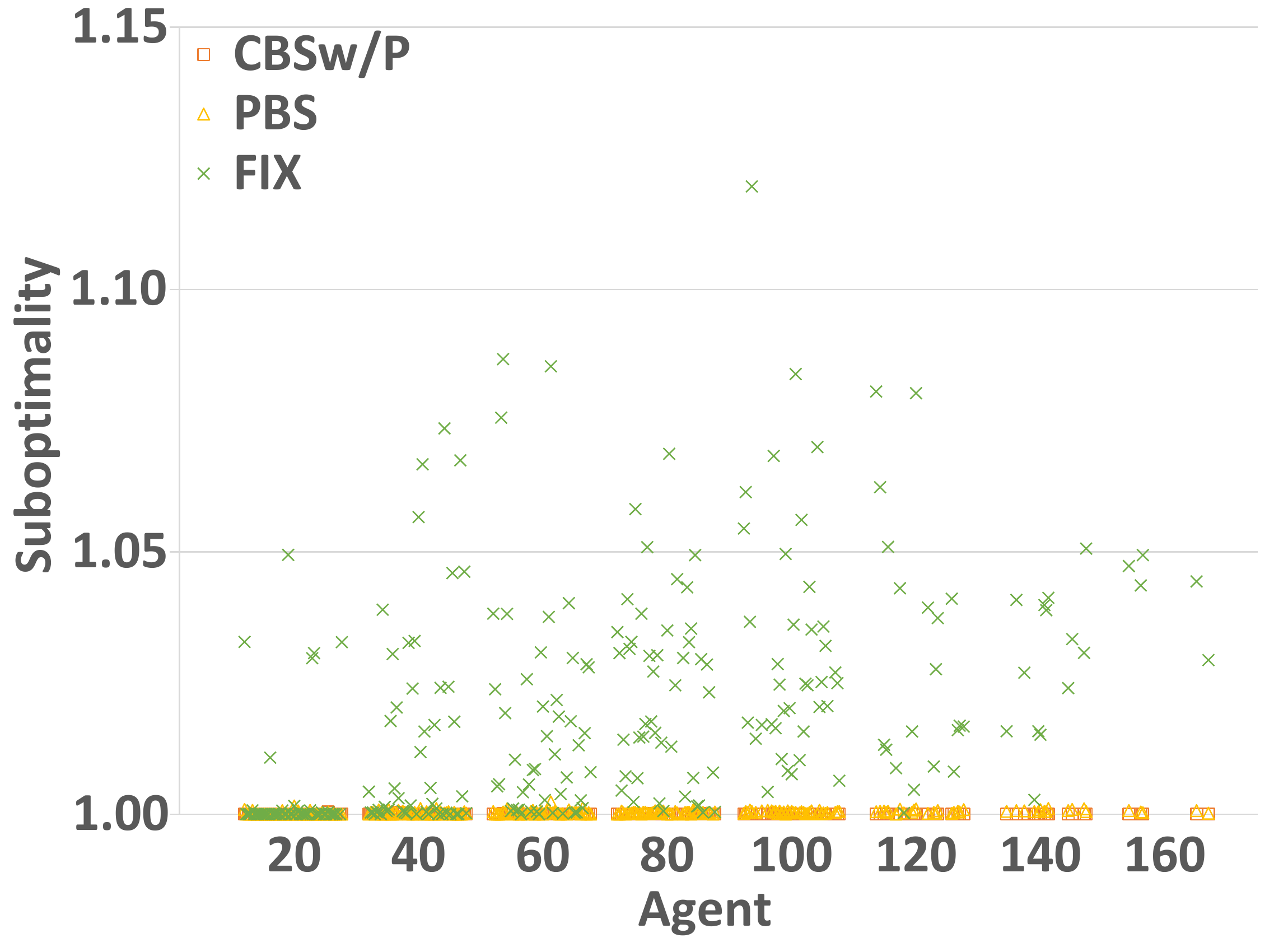}
  \caption{Suboptimality ratios for brc202d.}
\end{subfigure}%
\begin{subfigure}[b]{.5\columnwidth}
  \includegraphics[width=\columnwidth]{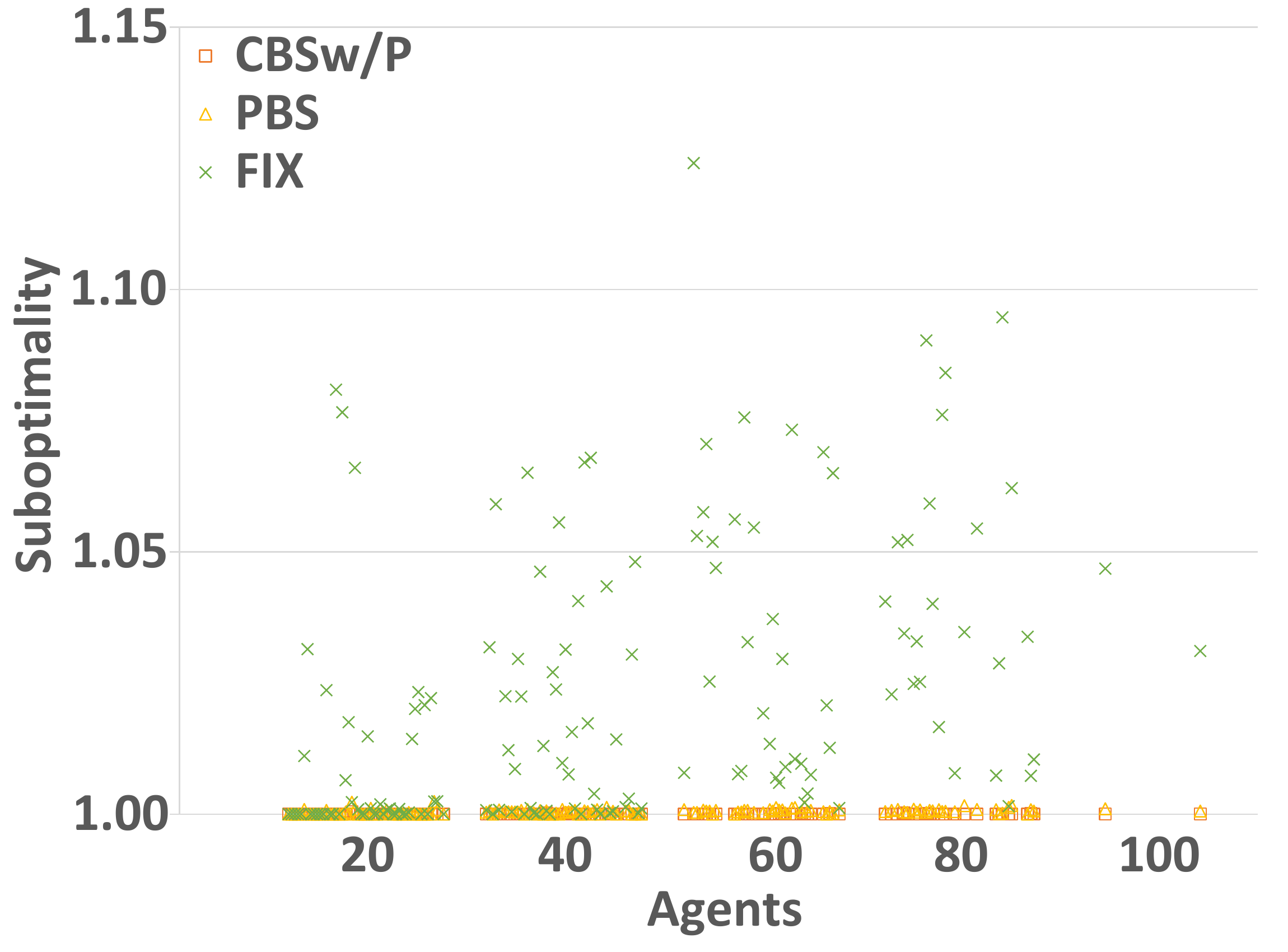}
  \caption{Suboptimality ratios for lak503d.}
\end{subfigure}\\
\begin{subfigure}[b]{.5\columnwidth}
\centering
  \includegraphics[width=\columnwidth]{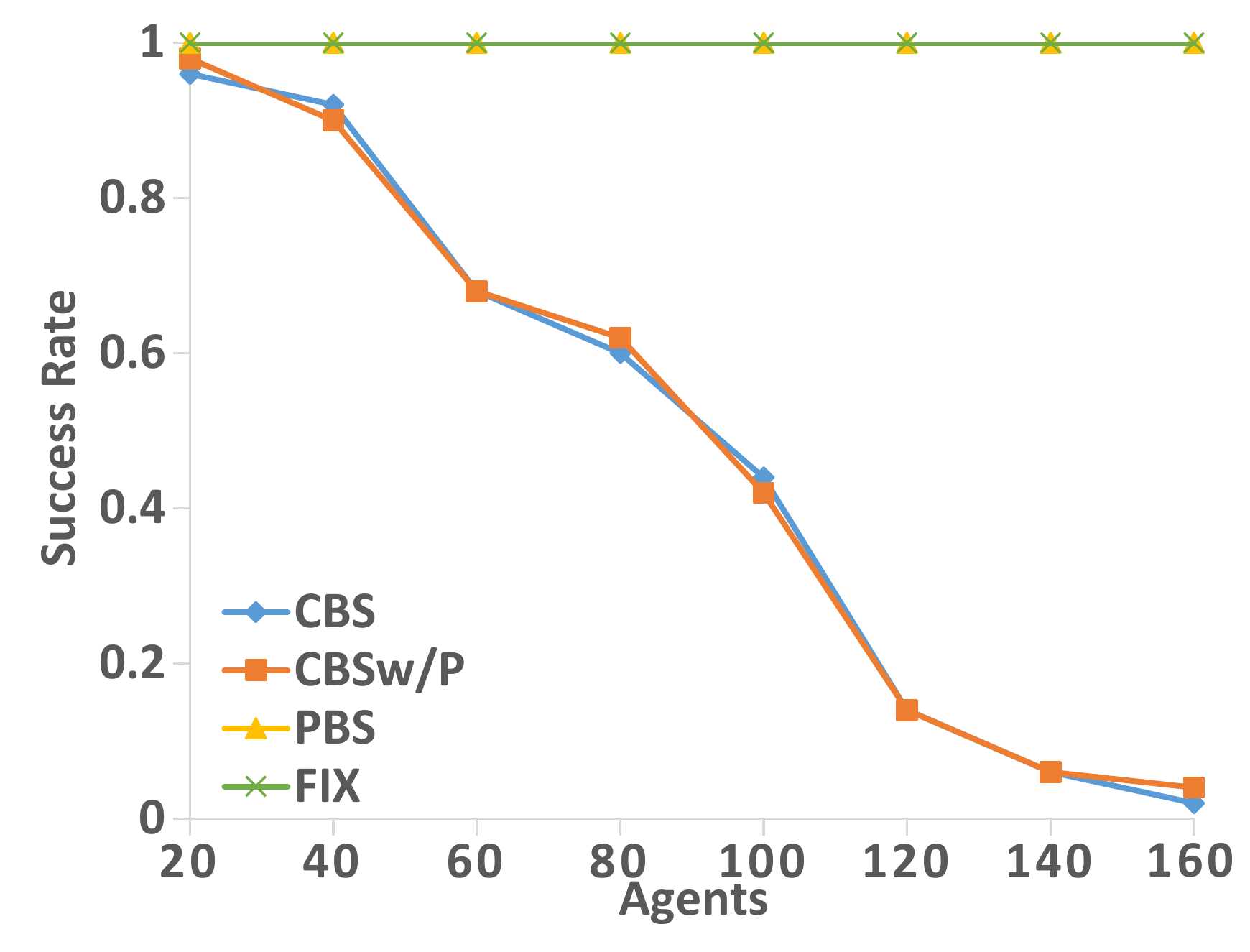}
  \caption{Success rates for brc202d(WF).}
\end{subfigure}%
\begin{subfigure}[b]{.5\columnwidth}
\centering
  \includegraphics[width=\columnwidth]{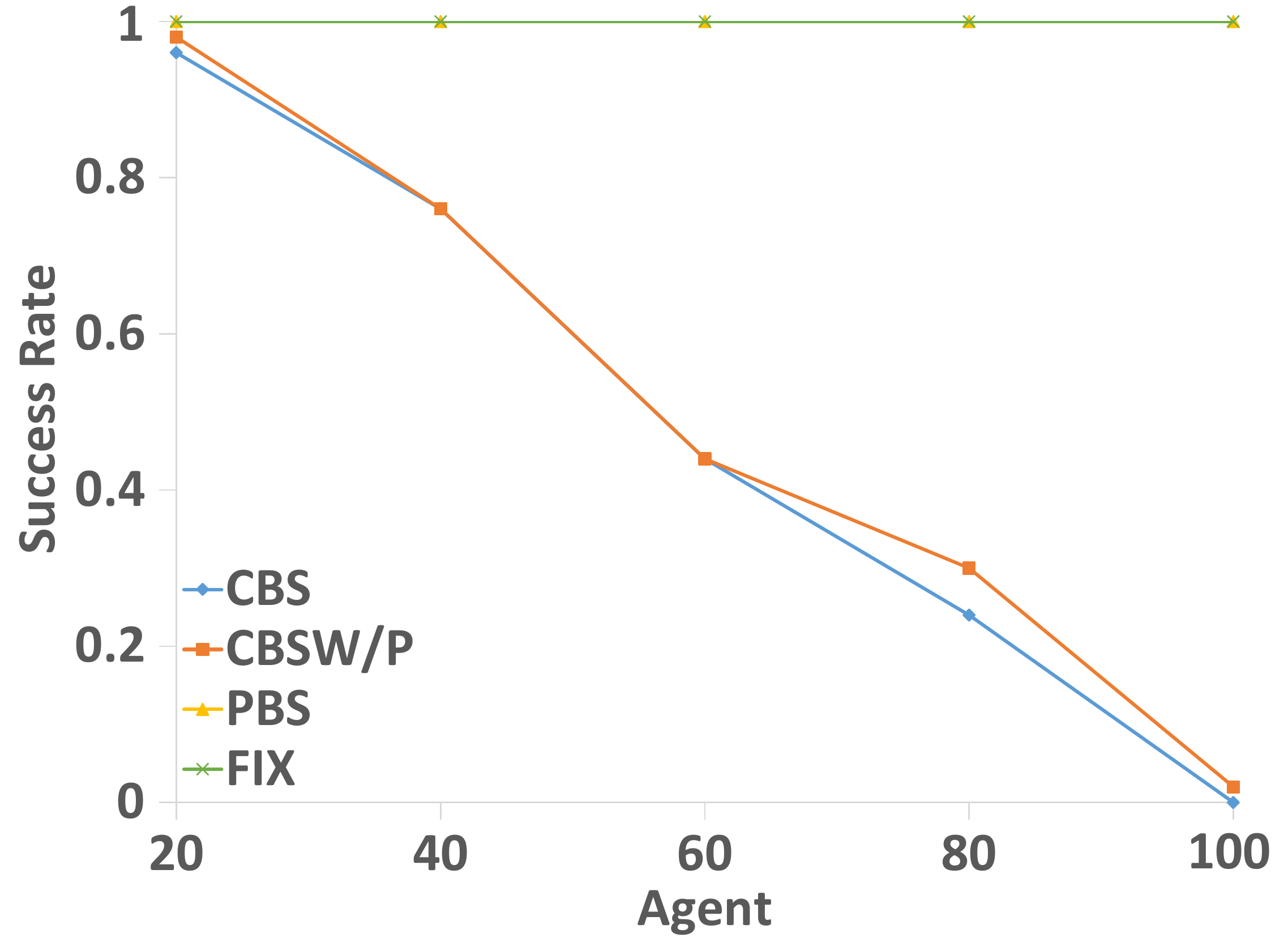}
  \caption{Success rates for lak503d(WF).}
\end{subfigure}\\
\begin{subfigure}[b]{.5\columnwidth}
  \includegraphics[width=\columnwidth]{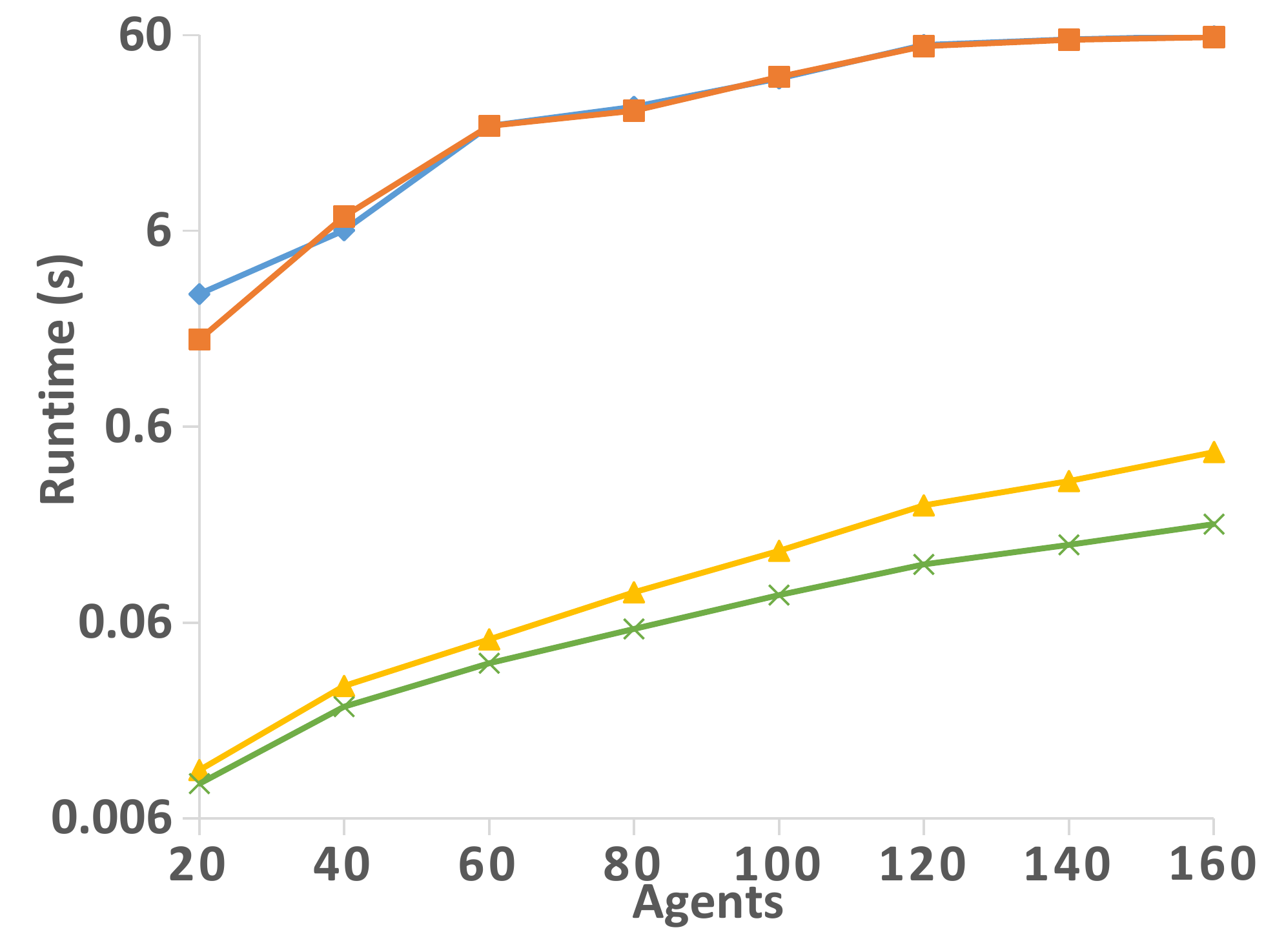}
  \caption{Runtimes for brc202d(WF).}
\end{subfigure}%
\begin{subfigure}[b]{.5\columnwidth}
  \includegraphics[width=\columnwidth]{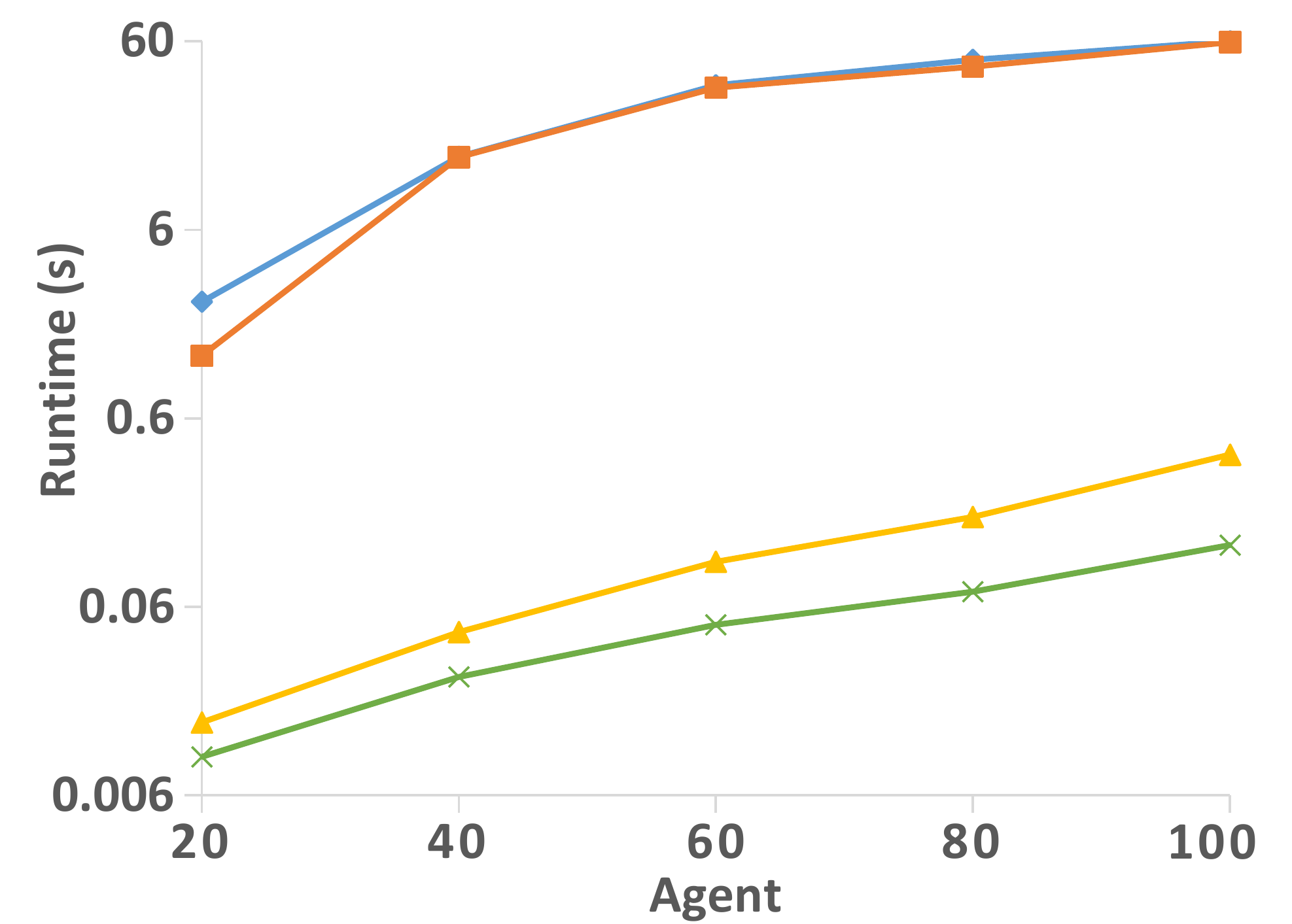}
  \caption{Runtimes for lak503d(WF).}
\end{subfigure}\\
\begin{subfigure}[b]{\columnwidth}
\renewcommand\arraystretch{0.6}
\setlength{\tabcolsep}{.75pt}
  %\bf\tiny
  \Large
\resizebox{\columnwidth}{!}{
\begin{tabular}{r|r|r|r|r|r|r|r|r|r|r|r|r|r}
\hline
\multicolumn{1}{c|}{}  & \multicolumn{1}{c|}{$M$} & \multicolumn{1}{c|}{sol} & \multicolumn{4}{c|}{\begin{tabular}[c]{@{\hspace{-\tabcolsep}}c@{\hspace{-\tabcolsep}}}low-level node expansions\\(\bm{$\times 1000$})\end{tabular}}                                                                                                       & \multicolumn{3}{c|}{\begin{tabular}[c]{@{\hspace{-\tabcolsep}}c@{\hspace{-\tabcolsep}}}high-level node\\expansions\end{tabular}}                                                                                 & \multicolumn{4}{c}{flowtime}                                                                                                                            \\
\hline
\multicolumn{1}{c|}{}     & \multicolumn{1}{c|}{}    & \multicolumn{1}{c|}{}    & \multicolumn{1}{c}{CBS} & \multicolumn{1}{c}{\begin{tabular}[c]{@{}c@{}}CBS\\ w/P\end{tabular}} & \multicolumn{1}{c}{PBS} & \multicolumn{1}{c|}{FIX} & \multicolumn{1}{c}{CBS} & \multicolumn{1}{c}{\begin{tabular}[c]{@{}c@{}}CBS\\ w/P\end{tabular}} & \multicolumn{1}{c|}{PBS} & \multicolumn{1}{c}{CBS} & \multicolumn{1}{c}{\begin{tabular}[c]{@{}c@{}}CBS\\ w/P\end{tabular}} & \multicolumn{1}{c}{PBS} & \multicolumn{1}{c}{FIX} \\
\hline
\parbox[t]{4mm}{\multirow{8}{*}{\rotatebox[origin=c]{90}{brc202d}}} & 20  & 49 & 3.33     & 3.18     & 15.64    & 15.24    & 0.55     & 0.55   & 0.45  & 2,529.16  & 2,529.18  & 2,529.49  & 2,538.55  \\
                         & 40  & 49 & 9.36     & 8.22     & 245.53   & 233.20   & 4.49     & 3.18   & 2.10  & 5,170.65  & 5,170.71  & 5,171.96  & 5,243.49  \\
                         & 60  & 47 & 16.13    & 15.37    & 341.22   & 381.73   & 6.60     & 6.64   & 4.04  & 7,517.43  & 7,517.45  & 7,519.79  & 7,647.45  \\
                         & 80  & 44 & 56.35    & 28.02    & 778.00   & 706.31   & 27.93    & 15.73  & 6.70  & 9,861.70  & 9,861.75  & 9,864.59  & 10,090.64 \\
                         & 100 & 37 & 87.41    & 59.50    & 1,294.59 & 1,326.70 & 39.70    & 30.92  & 10.03 & 12,344.73 & 12,344.78 & 12,349.43 & 12,740.68 \\
                         & 120 & 20 & 195.05   & 86.61    & 1,434.54 & 1,708.10 & 112.15   & 56.95  & 14.10 & 15,383.25 & 15,383.30 & 15,390.55 & 15,830.15 \\
                         & 140 & 13 & 388.29   & 159.20   & 1,402.82 & 1,302.98 & 114.77   & 61.69  & 18.77 & 16,709.92 & 16,710.00 & 16,721.38 & 17,194.23 \\
                         & 160 & 5  & 486.98   & 120.13   & 3,025.44 & 2,469.80 & 228.20   & 99.80  & 24.40 & 19,982.60 & 19,982.80 & 19,991.80 & 20,841.40 \\
                         \hline
\parbox[t]{4mm}{\multirow{5}{*}{\rotatebox[origin=c]{90}{lak503d}}} & 20  & 49 & 8.09     & 7.01     & 87.06    & 38.49    & 2.96     & 3.69   & 1.47  & 2,320.71  & 2,320.73  & 2,321.27  & 2,341.14  \\
                         & 40  & 46 & 49.12    & 22.04    & 232.75   & 197.86   & 22.26    & 11.24  & 4.20  & 4,702.54  & 4,702.59  & 4,704.35  & 4,785.83  \\
                         & 60  & 32 & 639.95   & 115.09   & 536.72   & 516.34   & 325.91   & 70.03  & 10.03 & 6,809.72  & 6,809.81  & 6,813.78  & 7,030.59  \\
                         & 80  & 25 & 783.31   & 360.23   & 844.29   & 905.59   & 341.28   & 251.20 & 15.00 & 9,121.60  & 9,121.76  & 9,127.88  & 9,485.76  \\
                         & 100 & 2  & 4,129.21 & 1,190.30 & 1,497.35 & 1,708.58 & 1,321.50 & 810.50 & 22.50 & 12,781.50 & 12,782.50 & 12,792.00 & 13,277.00  \\
\hline
\parbox[t]{6mm}{\multirow{7}{*}{\rotatebox[origin=c]{90}{\begin{tabular}[c]{@{}c@{}}brc202d\\ (WF)\end{tabular}}}} & 20  & 48  & 3.31                      & 3.16   & 3.07  & 2.74  & 0.50                  & 0.50   & 0.31 & 2,528.23  & 2,528.25  & 2,528.60  & 2,528.60  \\
                         & 40  & 45  & 9.31                      & 7.98   & 7.12  & 5.93  & 4.56                  & 3.02   & 1.22 & 5,179.53  & 5,179.56  & 5,180.82  & 5,180.82  \\
                         & 60  & 34  & 14.29                     & 13.19  & 10.67 & 8.85  & 5.88                  & 5.71   & 2.06 & 7,484.32  & 7,484.32  & 7,486.35  & 7,486.65  \\
                         & 80  & 30  & 20.80                     & 17.49  & 14.21 & 11.49 & 9.80                  & 8.87   & 3.17 & 9,890.00  & 9,890.00  & 9,893.03  & 9,893.30  \\
                         & 100 & 21  & 45.64                     & 31.13  & 20.63 & 15.48 & 23.52                 & 17.19  & 5.10 & 12,246.86 & 12,246.86 & 12,251.90 & 12,252.24 \\
                         & 120 & 7   & 133.41                    & 56.11  & 27.12 & 19.05 & 76.14                 & 35.29  & 7.57 & 15,079.71 & 15,079.71 & 15,087.29 & 15,087.43 \\
                         & 140 & 3   & 105.78                    & 68.06  & 31.95 & 22.61 & 57.00                 & 37.00  & 9.67 & 17,215.67 & 17,215.67 & 17,224.33 & 17,225.33 \\
                         \hline
\parbox[t]{6mm}{\multirow{4}{*}{\rotatebox[origin=c]{90}{\begin{tabular}[c]{@{}c@{}}lak503d\\ (WF)\end{tabular}}}} & 20  & 48  & 7.15                      & 6.37   & 4.45  & 3.11  & 2.54                  & 3.42   & 1.00 & 2,324.54  & 2,324.56  & 2,325.06  & 2,325.21  \\
                         & 40  & 37  & 33.14                     & 17.09  & 10.13 & 6.71  & 15.03                 & 8.41   & 2.59 & 4,648.08  & 4,648.08  & 4,649.68  & 4,650.27  \\
                         & 60  & 20  & 321.00                    & 89.70  & 19.90 & 11.44 & 117.05                & 41.55  & 6.10 & 6,744.40  & 6,744.40  & 6,748.50  & 6,749.30  \\
                         & 80  & 12  & 580.91                    & 167.59 & 28.39 & 14.83 & 258.00                & 84.83  & 9.08 & 9,033.58  & 9,033.58  & 9,040.58  & 9,041.17 \\
\hline
\end{tabular}
}
\caption{Results on instances solved by all CBS, CBSw/P, PBS, and FIX.}
\end{subfigure}
\caption{Results on game maps.\label{fig:game}}
\end{figure}

\noindent\textbf{Experiment 1: $\bm{20\times20}$ grids.}
We use MAPF instances on $20\times20$ four-neighbor grids with 0 and 10 percent obstacles (represented by randomly blocked cells) with random start and target vertices, in an empty grid (0\% obstacles) and a grid with 10\% of the cells filled with obstacles.
%The results are shown in Figure~\ref{fig:2020}.

Figures \ref{fig:2020}\subref{fig:succ_20} and \subref{fig:succ_20_obs} show the success rates of the algorithms, i.e., the percentage of MAPF instances solved within the time limit (RND solves a MAPF instance if it solves the MAPF instance in any of its ten runs within the time limit), as a function of the number of agents. The success rate of CBSw/P is higher than the one of CBS, although they both approach zero as the number of agents increases, due to the increasing size of the CT. PBS and its variants have much higher success rates. FIX, LH, and SH often fail to find a solution but PBS and RND find solutions for all MAPF instances except some ``hard'' ones with more than 70 agents and 10\% obstacles. For the ``hard'' instances, PBS never reports ``No Solution'' but reaches the runtime limit on some of them, while RND never reaches the runtime limit for all ten runs but reports ``No Solution'' for all ten runs on some of them. SH has the lowest success rates among all PBS variants with a total priority ordering because, in SH, the higher priority agents tend to arrive at their target vertices earlier and are thus more likely to block the lower priority agents from reaching their target vertices.
Figures~\ref{fig:2020}\subref{fig:time_20} and \subref{fig:time_20_obs} show the runtimes of the algorithms. The one minute runtime is averaged in for MAPF instances where the runtime limit was reached. CBSw/P is faster than CBS. RND is slower than the other PBS variants because it solves one MAPF instances multiple times. PBS is almost as fast as the other PBS variants with total priority orderings (within an order of magnitude always) but the runtime difference increases as the number of agents grows.

We then consider results for all MAPF instances that CBS, CBSw/P, PBS, and FIX solve. Figures~\ref{fig:2020}\subref{fig:sub_20} and \subref{fig:sub_20_obs} show the ratio of flowtime to the optimal flowtime (computed by CBS). We see that CBSw/P almost always finds optimal solutions. PBS finds solutions that are very close to optimal, getting slightly worse as the number of agents grows, but never more than 4\% worse than optimal. In contrast, FIX finds solutions that are more than 5\% worse than optimal and often much worse. The results in the graphs are confirmed by the results reported in the table shown in Figure~\ref{tab:2020_top} where the number of instances (\textbf{sol}) that we consider for each number of agents is also reported.

We also consider results for the set of instances that are solved by PBS, FIX, LH, SH, and RND simultaneously, so we can easily compare their behaviour. Figures~\ref{fig:2020}\subref{fig:rat_20} and \subref{fig:rat_20_obs} show that PBS is almost indistinguishable from SH, which is better than RND, which is in turn better than FIX, which is in turn better than LH. Figure~\ref{tab:2020_bot} shows that LH has the largest numbers of low-level node expansions among algorithms with total priority orderings. PBS has large numbers of low-level node expansions because it plans paths for more priority orderings and thus performs more low-level searches. RND has the largest numbers of low-level node expansions because it solves each MAPF instance multiple times. The runtime in Figures~\ref{fig:2020}\subref{fig:time_20} and \subref{fig:time_20_obs} behaves like the number of low-level node expansions since the algorithms spend most of their runtime in low-level searches.

\noindent\textbf{Experiment 2: Game Maps.} We also use MAPF instances on two standard benchmark maps \textbf{brc202d} (a $481\times 530$ four-neighbor grid) and \textbf{lak503d} (a $192\times 192$ four-neighbor grid) of the game Dragon Age: Origins \cite{sturtevant2012benchmarks}. We also use MAPF instances on these grids instances labeled \textbf{brc202d(WF)} and \textbf{lak503d(WF)} where the agents are removed from the grid once they arrive at their target vertices for the first time so that they cannot block other agents afterward. %The results are shown in Figure~\ref{fig:game}.

We first focus on results for brc202d and lak503d. CBSw/P outperforms CBS in terms of success rate (Figures~\ref{fig:game}(a) and (b)) and running time (Figures~\ref{fig:game}(c) and (d)), again with almost negligible reduction in solution quality (Figures~\ref{fig:game}(e) and (f)).
PBS outperforms FIX, CBS, and CBSw/P in terms of success rate, and is close to FIX in terms of runtime.
PBS is almost always able to find nearly optimal solutions as opposed to FIX (Figures~\ref{fig:game}(e) and (f)).
Figures~\ref{fig:game}(k) shows results for all MAPF instances that all algorithms solve within the runtime limit. PBS and FIX expand more low-level nodes than CBS and CBSw/P for \textbf{brc202d} because the narrow corridors make the constraints of high priority agents more significant than for other maps, especially when those agents arrive at their target vertices early and stop moving, which explains why PBS and FIX reach the runtime limit for some instances for 120 and 160 agents on \textbf{brc202d} within the time limit.

We then focus on results for brc202d(WF) and lak503d(WF). These MAPF instances are much easier to solve for PBS and FIX than those on brc202d and lak503d, as the numbers of low-level node expansions show. CBSw/P and CBS are almost indistinguishable with respect to both success rates and runtimes (Figures~\ref{fig:game}(g) to (j)), which is not the case for brc202d and lak503d.

We further tested the scalability of PBS on brc202d(WF) with large numbers of agents. PBS scales to 600 agents and never reaches the runtime limit (shown in the below table).
\begin{center}\Large
\resizebox{0.8\columnwidth}{!}{
\begin{tabular}{c|r|r|r|r|r|r}
\hline
$M$                & 100  & 200   & 300   & 400   & 500    & 600    \\\hline
runtime (s)            & 0.14 & 0.86  & 3.07  & 8.79  & 18.51  & 35.18  \\\hline
PT node expansions & 5.16 & 21.38 & 46.02 & 81.84 & 130.04 & 182.74 \\\hline
\end{tabular}
}\end{center}

\section{Conclusions}

We tackled the main challenge of exploring ``good'' orderings of agents for prioritized planning from a conceptual and practical perspective. From the conceptual side, we developed a first theoretical framework for discussing the limits of prioritized planning. From the practical side, we developed two new algorithms, CBSw/P and PBS, that search for ``good'' orderings and thus compute solutions with ``good'' orderings. Both algorithms order agents lazily by imposing an ordering on two agents only to resolve collisions between them and both of them explore the orderings in a systematic fashion: CBSw/P using best-first search, and PBS using depth-first search.

\small
\bibliographystyle{aaai}
\bibliography{references}

\end{document}